\documentclass[10pt,twocolumn,letterpaper]{article}

\usepackage{caption}   

\usepackage{cvpr}
\usepackage{times}
\usepackage{epsfig}
\usepackage{graphicx}
\usepackage{amsmath,amssymb} 
\usepackage{amssymb}

\usepackage{comment}
\usepackage{array}
\usepackage{cellspace}
\usepackage{multirow}
\usepackage{makecell}

\usepackage{fixltx2e}
\usepackage{float}
\usepackage{dblfloatfix}

\usepackage{booktabs}
\usepackage[table, x11names]{xcolor}
\usepackage[normalem]{ulem}
\usepackage{hhline}
\usepackage{placeins}

\usepackage{enumitem}

\newcolumntype{P}[1]{>{\centering\arraybackslash}p{#1}}

\newcommand{\Gbox}{G_\text{box}}
\newcommand{\Gmask}{G_\text{mask}}
\newcommand{\Gimg}{G_\text{img}}

\newcommand{\txt}{\mathbf{s}}
\newcommand{\bbox}{\mathbf{b}}
\newcommand{\B}{B}
\newcommand{\cls}{{\boldsymbol l}}

\newcommand{\mask}{M}

\newcommand{\inst}{\text{inst}}
\newcommand{\glob}{\text{global}}

\newcommand{\Dimg}{D_{\text{img}}}

\newcommand{\img}{X}

\makeatletter
\renewcommand{\paragraph}{%
  \@startsection{paragraph}{4}%
  {\z@}{1.0ex \@plus 0.50ex \@minus 0.20ex}{-1em}%
  {\normalfont\normalsize\bfseries}%
}
\makeatother

\iftrue 
        \newcommand{\cutsectionup}{\vspace*{-0.07in}}
        \newcommand{\cutsectiondown}{\vspace*{-0.05in}}

        \newcommand{\cutsubsectiondown}{\vspace*{-0.04in}}

        \newcommand{\cutparagraphup}{\vspace*{-0.02in}}

        \newcommand{\cutabstractup}{\vspace*{-0.12in}}
        \newcommand{\cutabstractdown}{\vspace*{-0.13in}}

\else 
        \newcommand{\cutsectionup}{}
        \newcommand{\cutsectiondown}{}

        \newcommand{\cutsubsectiondown}{}

        \newcommand{\cutparagraphup}{}

        \newcommand{\cutabstractup}{}
        \newcommand{\cutabstractdown}{}

\fi

\usepackage[pagebackref=true,breaklinks=true,letterpaper=true,colorlinks,bookmarks=false]{hyperref}

\cvprfinalcopy 

\def\cvprPaperID{1696} 

\ifcvprfinal\fi
\begin{document}

\title{Inferring Semantic Layout for Hierarchical Text-to-Image Synthesis}


\author{
Seunghoon Hong\textsuperscript{$^\dagger$}\hspace{0.75cm} Dingdong Yang\textsuperscript{$^\dagger$}\hspace{0.75cm} Jongwook Choi\textsuperscript{$^\dagger$}\hspace{0.75cm} Honglak Lee\textsuperscript{$^{\ddagger,\dagger}$}
\smallskip 
\\
\textsuperscript{$\dagger$}University of Michigan \\
\textsuperscript{$\ddagger$}Google Brain \\
\begin{tabular}{c c}
\textsuperscript{$\dagger$}{\tt\small \{hongseu,didoyang,jwook,honglak\}@umich.edu} & 
\textsuperscript{$\ddagger$}{\tt\small honglak@google.com} \\
\end{tabular}
%
\vspace*{-3pt}
}

\maketitle

\begin{abstract}
\cutabstractup
We propose a novel hierarchical approach for text-to-image synthesis by inferring semantic layout.
%
%
%
%
%
%
Instead of learning a direct mapping from text to image, our algorithm decomposes the generation process into multiple steps, in which it first constructs a semantic layout from the text by the layout generator and converts the layout to an image by the image generator.
The proposed layout generator progressively constructs a semantic layout in a coarse-to-fine manner by generating object bounding boxes and refining each box by estimating object shapes inside the box. 
The image generator synthesizes an image conditioned on the inferred semantic layout, which provides a useful semantic structure of an image matching with the text description. 
Our model not only generates semantically more meaningful images, but also allows automatic annotation of generated images and user-controlled generation process by modifying the generated scene layout. 
We demonstrate the capability of the proposed model on challenging MS-COCO dataset and show that the model can substantially improve the image quality, interpretability of output and semantic alignment to input text over existing approaches.
\cutabstractdown
\end{abstract}

\cutsectionup
\section{Introduction}
\label{sec:introduction}
\cutsectiondown

\ifdefined\paratitle {\color{blue}
[conditional GAN is effective in generating images in specific context, but not for complex context] \\
} \fi
Generating images from text description has been an active research topic in computer vision. 
By allowing users to describe visual concepts in natural language, it provides a natural and flexible interface for conditioning image generation.  
Recently, approaches based on conditional Generative Adversarial Network (GAN) have shown promising results on text-to-image synthesis task~\cite{ReedS16ICML,HanZ17,ReedS16Nips}.
Conditioning both generator and discriminator on text,
these approaches are able to generate realistic images that are both diverse and relevant to input text.
Based on conditional GAN framework, recent approaches further improve the prediction quality by generating high-resolution images~\cite{HanZ17} or augmenting text information~\cite{hao17ICIP,dash17Arxiv}.

\begin{figure}[!t]
\centering
\includegraphics[width=0.93\columnwidth]{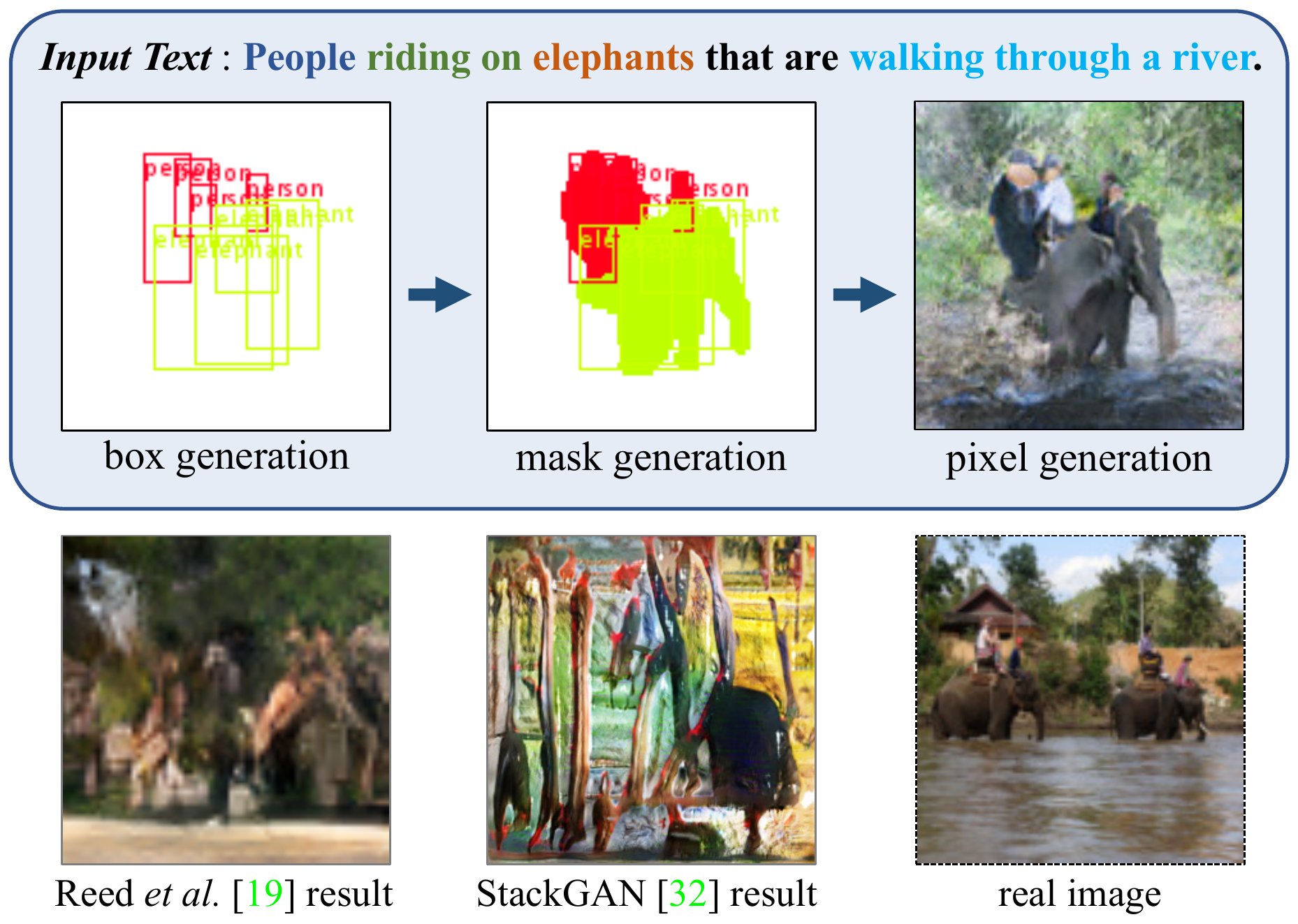}
\caption{
Overall framework of the proposed algorithm. 
Given a text description, our algorithm sequentially constructs a semantic structure of a scene and generates an image conditioned on the inferred layout and text.
Best viewed in color.
}
\label{fig:overview}
\vspace{-0.45cm}
\end{figure}

However, the success of existing approaches has been mainly limited to simple datasets such as birds~\cite{WelinderEtal2010} and flowers~\cite{Nilsback08}, while generation of complicated, real-world images such as MS-COCO~\cite{Lin:2014:MSCOCO}  remains an open challenge.
As illustrated in Figure~\ref{fig:overview}, generating image from a general sentence \emph{``people riding on elephants that are walking through a river''} requires multiple reasonings on various visual concepts, 
such as object category (\emph{people} and \emph{elephants}),
spatial configurations of objects (\emph{riding}),
scene context (\emph{walking through a river}), \emph{etc.},
which is much more complicated than generating a single, large object as in simpler datasets~\cite{WelinderEtal2010, Nilsback08}.
Existing approaches have not been successful in generating reasonable images for such complex text descriptions,
because of the complexity of learning a direct text-to-pixel mapping from general images.

\ifdefined\paratitle {\color{blue}
[Layout generator can generate interpretable images even in complex image.] \\
} \fi
Instead of learning a direct mapping from text to image, we propose an alternative approach that constructs \textit{semantic layout} as an intermediate representation between text and image. 
Semantic layout defines a structure of scene based on object instances and provides fine-grained information of the scene, such as the number of objects, object category, location, size, shape, \etc (Figure~\ref{fig:overview}).    
By introducing a mechanism that explicitly aligns the semantic structure of an image to text, the proposed method can generate complicated images that match complex text descriptions.
In addition, conditioning the image generation on semantic structure allows our model to generate semantically more meaningful images that are easy to recognize and interpret.

\ifdefined\paratitle {\color{blue}
[Our approach bridges the gap between text-to-image synthesis and layout-conditional generation] \\
} \fi
Our model for hierarchical text-to-image synthesis consists of two parts:
%
the \textit{layout generator} that constructs a semantic label map from a text description,
and the \textit{image generator} that converts the estimated layout to an image using the text. 
Since learning a direct mapping from text to fine-grained semantic layout is still challenging,
we further decompose the task into two manageable subtasks:
we first estimate the bounding box layout of an image using the \emph{box generator},
and then refine the shape of each object inside the box by the \emph{shape generator}.
The generated layout is then used to guide the image generator for pixel-level synthesis.
The box generator, shape generator and image generator are implemented by independent neural networks, 
and trained in parallel with corresponding supervisions. 

\ifdefined\paratitle {\color{blue}
[Benefits of exploiting intermediate layout structure] \\
} \fi
Generating semantic layout not only improves quality of text-to-image synthesis, but also provides a number of potential benefits.
First, the semantic layout provides instance-wise annotations on generated images,
which can be directly exploited for automated scene parsing and object retrieval.
Second, it offers an interactive interface for controlling image generation process; users can modify the semantic layout to generate a desired image by removing/adding objects, changing size and location of objects, \etc.
The contributions of this paper are as follows:
\begin{itemize}[topsep=3pt, parsep=3pt, itemsep=0pt]
    \item
    We propose a novel approach for synthesizing images from complicated text descriptions.
    Our model explicitly constructs semantic layout from the text description, and guides image generation using the inferred semantic layout. 

    \item
    By conditioning image generation on
    explicit layout prediction, our method is able to generate images that are semantically meaningful and well-aligned with input descriptions. 
    \item
    We conduct extensive quantitative and qualitative evaluations on challenging MS-COCO dataset,
    and demonstrate substantial improvement on generation quality over existing works. 
\end{itemize}

The rest of the paper is organized as follows.
We briefly review related work in Section~\ref{sec:relatedwork}, and provide an overview of the proposed approach in Section~\ref{sec:overview}.
Our model for layout and image generation is introduced in Section~\ref{sec:layout_generator} and \ref{sec:pixel_gen}, respectively.
We discuss the experimental results on the MS-COCO dataset in Section~\ref{sec:experiment}.

\cutsectionup
\section{Related Work}
\label{sec:relatedwork}
\cutsectiondown

\ifdefined\paratitle {\color{blue} 
[Text to image synthesis] \\
} \fi
Generating images from text descriptions has recently drawn a lot of attention from the research community.
Formulating the task as a conditional image generation problem,
various approaches have been proposed based on 
Variational Auto-Encoders (VAE)~\cite{Mansimov16ICLR},
auto-regressive models~\cite{ReedS17ICML},
optimization techniques~\cite{nguyen17CVPR}, \textit{etc}. 
Recently, approaches based on conditional Generative Adversarial Network (GAN) \cite{Goodfellow:NIPS2014:GAN} have shown promising results in
text-to-image synthesis~\cite{ReedS16ICML, ReedS16Nips, HanZ17, hao17ICIP, dash17Arxiv}.
Reed \etal~\cite{ReedS16ICML} proposed to learn both generator and discriminator conditioned on text embedding.
Zhang \etal~\cite{HanZ17} improved the image quality by increasing image resolution with a two-stage GAN.
Other approaches include improving conditional generation by augmenting text data with synthesized captions~\cite{hao17ICIP},
or adding conditions on class labels~\cite{dash17Arxiv}.
Although these approaches have demonstrated impressive generation results on
datasets of specific categories (\eg, birds~\cite{WelinderEtal2010} and flowers~\cite{Nilsback08}), 
the perceptual quality of generation tends to substantially degrade on datasets with complicated images (\eg, MS-COCO~\cite{Lin:2014:MSCOCO}). 
We investigate a way to improve text-to-image synthesis on general images,
by conditioning generation on inferred semantic layout.

\ifdefined\paratitle {\color{blue} 
[Conditional generation on image structure] \\
} \fi
The problem of generating images from pixel-wise semantic labels has been explored recently~\cite{ChenQ17,Isola17CVPR,KaracanAEE16,ReedS17ICML}. 
In these approaches, the task of image generation is formulated as translating semantic labels to pixels.
Isola \etal~\cite{Isola17CVPR} proposed a pixel-to-pixel translation network that converts dense pixel-wise labels to image, and  
Chen \etal~\cite{ChenQ17} proposed a cascaded refinement network that generates high-resolution output from dense semantic labels. 
Karacan \etal~\cite{KaracanAEE16} employed both dense layout and attribute vectors for image generation using conditional GAN.
Notably, Reed \etal~\cite{ReedS17ICML} utilized sparse label maps like our method.
Unlike previous approaches that require ground-truth layouts for generation,
our method \textit{infers} the semantic layout,
and thus is more generally applicable to various generation tasks. 
Note that our main contribution is complementary to these approaches, and we can integrate existing segmentation-to-pixel generation methods to generate an image conditioned on a layout inferred by our method.

\ifdefined\paratitle {\color{blue} 
[Inferring structure] \\
} \fi
The idea of inferring scene structure for image generation is not new, as it has been explored by some recent works in several domains.
For example,
Wang \etal~\cite{WangX16ECCV} proposed to infer a surface normal map as an intermediate structure to generate indoor scene images,
and
Villegas \etal~\cite{Villegas17ICML} predicted human joints for future frame prediction. 
The most relevant 
work to our method is Reed \etal~\cite{ReedS16Nips},
which predicted local key-points of bird or human for text-to-image synthesis.
Contrary to the previous approaches that predict such specific types of structure for image generation,
our proposed method aims to predict semantic label maps, which is a general representation of natural images.
%
%

%
%

%
%

%
\begin{figure*}[!ht] \begin{center}
    \vspace{-0.3cm}
    \includegraphics[width=0.97\textwidth]{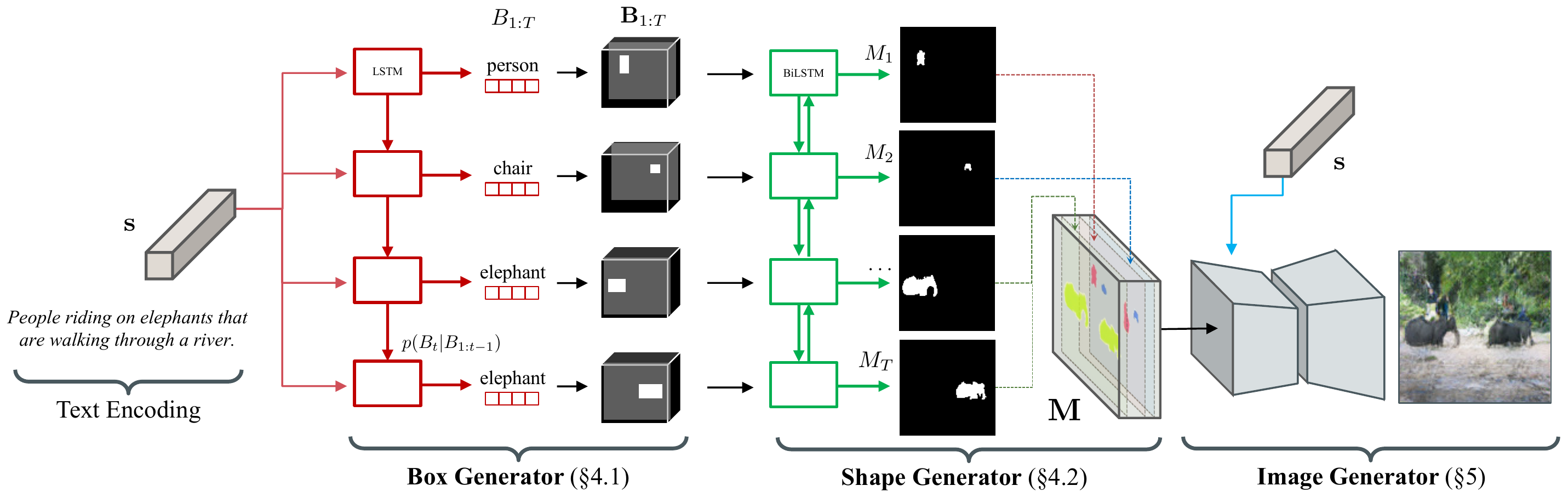}
    \caption{Overall pipeline of the proposed algorithm. 
    Given a text embedding, our algorithm first generates a coarse layout of the image by placing a set of object bounding boxes
    using the box generator (Section~\ref{sec:box_gen}), and further refines the object shape inside each box using the shape generator (Section~\ref{sec:mask_gen}). 
    Combining outputs from the box and the shape generator leads to a semantic label map defining semantic structure of the scene.
    Conditioned on the inferred semantic layout and the text,
    a pixel-wise image is finally generated by the image generator (Section~\ref{sec:pixel_gen}).      
    }
    \label{fig:architecture}
    \vspace{-0.6cm}
\end{center} \end{figure*}

\cutsectionup
\section{Overview}
\label{sec:overview}
\cutsectiondown

The overall pipeline of the proposed framework is illustrated in Figure~\ref{fig:architecture}.
Given a text description, our model progressively constructs a scene by refining semantic structure of an image using the following sequence of generators:

\begin{itemize}[topsep=2pt, parsep=3pt, itemsep=0pt]
    \item \textbf{Box generator} 
    takes a text embedding $\txt$ as input, and generates a coarse layout by composing object instances in an image. 
    The output of the box generator is a set of bounding boxes $\B_{1:T}=\{\B_1,...,\B_T\}$,
    where each bounding box $\B_t$ defines the location, size and category label of the $t$-th object (Section~\ref{sec:box_gen}).
    
    \item \textbf{Shape generator} 
    takes a set of bounding boxes generated from box generator, and predicts shapes of the object inside the boxes.
    The output of the shape generator is a set of binary masks $\mask_{1:T}=\{\mask_1,...,\mask_T\}$,
    where each mask $\mask_t$ defines the foreground shape of the $t$-th object (Section~\ref{sec:mask_gen}). 

    \item \textbf{Image generator} 
    takes the semantic label map $\mathbf M$ obtained by aggregating instance-wise masks, and the text embedding as inputs, 
    and generates an image by translating a semantic layout to pixels matching the text description (Section~\ref{sec:pixel_gen}).

\end{itemize}

By conditioning the image generation process on the semantic layouts
that are explicitly inferred,
our method is able to generate images that preserve detailed object shapes
    and therefore are easier to recognize semantic contents.
In our experiments, we show that the images generated by our method are semantically more meaningful
and well-aligned with the input text,
compared to ones generated by previous approaches \cite{ReedS16ICML,HanZ17} (Section~\ref{sec:experiment}). 
%

%
\cutsectionup
\section{Inferring Semantic Layout from Text}
\label{sec:layout_generator}
\cutsectiondown

%
%

%
%
%
%

\subsection{Bounding Box Generation}
\label{sec:box_gen}
\cutsubsectiondown

Given an input text embedding $\txt$,
we first generate a coarse layout of image in the form of object bounding boxes.
We associate each bounding box $B_t$ with a class label
to define which class of object to place and where,
which plays a critical role in determining the global layout of the scene.
Specifically, we denote the labeled bounding box of the $t$-th object as $B_t=(\bbox_t, \cls_t)$,
where $\bbox_t=[b_{t,x}, b_{t,y}, b_{t,w}, b_{t,h}] \in \mathbb{R}^4$ 
represents the location and size of the bounding box,
and $\cls_t \in \{0, 1\}^{L+1}$ is a one-hot class label over $L$ categories.
We reserve the $(L+1)$-th class as a special indicator for the end-of-sequence.

The \emph{box generator} $\Gbox$ defines a stochastic mapping
from the input text $\txt$ to
a set of $T$ object bounding boxes
$\B_{1:T} = \{B_1,...,B_T\}$: 
\begin{equation}
    \widehat{\B}_{1:T} \sim \Gbox(\txt).
    \label{eqn:box_generator}
\end{equation}
%
%
%
%
%
%
%
%
%

\paragraph{Model.}
We employ an auto-regressive decoder for the box generator,
by decomposing the conditional joint bounding box probability
as
$p(B_{1:T} \mid \txt) = \prod_{t=1}^{T} p(B_t \mid B_{1:t-1}, \txt)$,
where the conditionals are approximated by LSTM~\cite{HochreiterS97}.
In the generative process, we first sample a class label $\cls_t$ for the $t$-th object
and then generate the box coordinates $\bbox_t$ conditioned on $\cls_t$,
\ie, $p(B_t | \cdot) = p(\bbox_t, \cls_t | \cdot) = p(\cls_t | \cdot) \, p(\bbox_t | \cls_t, \cdot)$.
The two conditionals are modeled by a Gaussian Mixture Model (GMM)
and a categorical distribution~\cite{sketchrnn}, respectively:
\begin{align}
    \vspace{-5pt}
    p(\cls_t \mid \B_{1:t-1}, \txt)
        &= \mathrm{Softmax}(\mathbf e_t),
    \label{eqn:cls_output}
    \\
    p(\bbox_t \mid \cls_t,B_{1:t-1}, \txt)
        &=  \sum_{k=1}^K \pi_{t,k} \, \mathcal{N}\left( \bbox_t ; \boldsymbol\mu_{t,k}, \boldsymbol\Sigma_{t,k} \right),
    \label{eqn:box_output}
\end{align}
where $K$ is the number of mixture components.
The softmax logit $\mathbf e_t$ in Eq.\eqref{eqn:cls_output}
and the parameters for the Gaussian mixtures
$\pi_{t,k} \in \mathbb R, \boldsymbol \mu_{t,k} \in \mathbb R^4$ and $\boldsymbol\Sigma_{t, k} \in \mathbb R^{4 \times 4}$
in Eq.\eqref{eqn:box_output}
are computed by the outputs from each LSTM step $t$.
Please see Section~\ref{sec:box_detail} in the appendix for details.

%
%
%

%
%

\paragraph{Training.}
We train the box generator by minimizing
the negative log-likelihood of ground-truth bounding boxes:
\begin{align}
    \mathcal{L}_{\text{box}} =
    -\lambda_l \,\frac{1}{T} \sum_{t=1}^{T} \cls^*_{t} \log p(\cls_{t})
    -\lambda_b \,\frac{1}{T} \sum_{t=1}^{T} \log p(\bbox^*_{t}),
\label{eqn:obj_box_generator}
\end{align}
where
$T$ is the number of objects in an image,
and $\lambda_l, \lambda_b$ are balancing hyper-parameters,
which are set to 4 and 1 in our experiment, respectively.
$\bbox^*_{t}$ and $\cls^*_{t}$ are ground-truth bounding box coordinates and label of the $t$-th object, respectively,
which are ordered based on their bounding box locations from left to right.
Note that we drop the conditioning in Eq.~\eqref{eqn:obj_box_generator} for notational brevity.

At test time, we generate 
bounding boxes via ancestral sampling of
box coordinates and class label
by Eq.~\eqref{eqn:cls_output} and \eqref{eqn:box_output}, respectively.
We terminate the sampling when the sampled class label corresponds to the termination indicator $(L+1)$, thus the number of objects are determined adaptively based on the text.

\subsection{Shape Generation}
\label{sec:mask_gen}

Given a set of bounding boxes obtained by the box generator,
the shape generator predicts more detailed image structure in the form of object masks.
Specifically, for each object bounding box $B_t$ obtained by Eq.~\eqref{eqn:box_generator},
we generate a binary mask $\mask_t\in\mathbb{R}^{H\times W}$
that defines the shape of the object inside the box. 
To this end, we first convert the discrete bounding box outputs $\{B_t\}$
to a binary tensor $\mathbf B_t \in \{0,1\}^{H\times W \times L}$,
whose element is 1 if and only if it is contained in the corresponding class-labeled box.
Using the notation $\mask_{1:T}=\{M_1, ..., M_T\}$,
we define the \emph{shape generator} $\Gmask$ as
\begin{align}
    \widehat{\mask}_{1:T} = \Gmask(\mathbf B_{1:T}, \mathbf z_{1:T}),
    \label{eqn:mask_generator}
\end{align}
where
$\mathbf z_t \sim \mathcal N(0, I)$ is a random noise vector.

Generating an accurate object shape should meet two requirements:
(i) First, each instance-wise mask $\mask_t$ should match the location and class information of $\mathbf B_t$,
and be recognizable as an individual instance
({instance-wise constraints}).
(ii) Second, each object shape must be aligned with its surrounding context ({global constraints}).
To satisfy both, 
we design the shape generator as a recurrent neural network,
which is trained with two conditional adversarial losses as described below.

\paragraph{Model.}
%
%
We build the shape generator $\Gmask$ using a convolutional recurrent neural network~\cite{Shi15Nips}, as illustrated in Figure~\ref{fig:architecture}.
At each step $t$, the model takes $\mathbf B_t$ through encoder CNN, and encodes information of all object instances by bi-directional convolutional LSTM (Bi-convLSTM).
On top of convLSTM output at $t$-th step, we add noise $\mathbf z_t$ by spatial tiling and concatenation, and generate a mask $\mask_t$ by forwarding it through a decoder CNN.

%
%
%

%
%
%
%
%
%
%
%
%
%

%
%
%
%
%
%
%
%
%

%
%
%
%
%
%
%

%

%
%
%
%
%

%
%
%
%
%
%
%

\begin{figure*}[!t]
    \centering
    \vspace{-0.2cm}
    \includegraphics[width=0.93\textwidth]{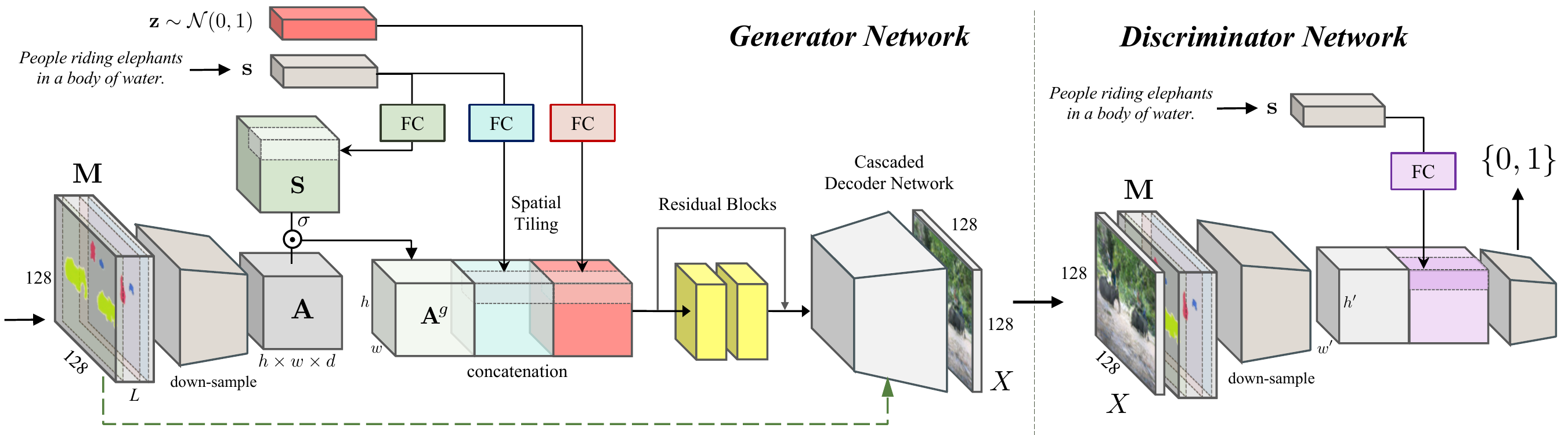}
    \vspace{-0.05cm}
    \caption{
    Architecture of the image generator.
    Conditioned on the text description and the semantic layout generated by the layout generator,
    it generates an image that matches both inputs.%
    }
    \label{fig:image_generator}
    \vspace{-0.3cm}
\end{figure*}

\paragraph{Training.}
Training of the shape generator is based on
the GAN framework \cite{Goodfellow:NIPS2014:GAN},
in which generator and discriminator are alternately trained.
To enforce both the global and the instance-wise constraints discussed earlier, 
we employ two conditional adversarial losses~\cite{Mirza:CGAN}
with the instance-wise discriminator $D_\inst$ and
the global discriminator $D_\glob$. 

First, we encourage each object mask to be compatible with class and location information encoded by object bounding box.
We train an instance-wise discriminator $D_\inst$
by optimizing the following \emph{instance-wise adversarial loss}:
\begin{align}
    \mathcal L_{\text{inst}}^{(t)} &=
        \mathbb E_{ (\mathbf{B}_t,\mask_t) } \Big[
            \log D_\inst \big( \mathbf{B}_t, \mask_t \big)
        \Big]
    \label{eqn:shape_instance_loss} \\
    &~~~ +
        \mathbb E_{\mathbf{B}_t, \mathbf{z}_t } \Big[
            \log \Big( 1 - D_\inst \big(\mathbf{B}_t, \Gmask^{(t)}(\mathbf{B}_{1:T}, \mathbf{z}_{1:T}) \big) \Big)
        \Big]
    , \nonumber
\end{align}
where $\Gmask^{(t)}(\mathbf{B}_{1:T}, \mathbf{z}_{1:T})$ indicates the $t$-th output from mask generator.
The instance-wise loss is applied for each of $T$ instance-wise masks, and aggregated over all instances as
$\mathcal{L}_\inst = (1/T) \sum_t \mathcal{L}_\inst^{(t)}$.

On the other hand, the global loss encourages all the instance-wise masks
form a globally coherent context.
To consider relation between different objects, we aggregate them into a global mask%
\footnote{$G_\text{global}$ is computed by addition to model overlap between objects.}
$
    G_\glob(\mathbf B_{1:T}, \mathbf z_{1:T}) = \textstyle \sum_t G_\text{mask}^{(t)}(\mathbf{B}_{1:t}, \mathbf z_{1:t}),
$
and compute an global adversarial loss analogous to Eq.~\eqref{eqn:shape_instance_loss} as
\begin{align}
    &\mathcal L_\glob =
        \mathbb E_{ (\mathbf{B}_{1:T},\mask_{1:T}) } \Big[
            \log D_\glob \big( \mathbf{B}_\glob, \mask_\glob \big)
        \Big]
    \label{eqn:shape_global_loss} \\
    &~~ +
        \mathbb E_{\mathbf{B}_{1:T}, \mathbf{z}_{1:T} } \Big[
            \log \hspace{-0.2em}\Big(\hspace{-0.1em}
                1 - D_\glob \big(
                    \mathbf{B}_\glob, G_\glob(\mathbf{B}_{1:T}, \mathbf{z}_{1:T}) \big)
            \hspace{-0.2em}\Big)
        \Big]
    , \nonumber
\end{align}
where $\mask_\glob\in\mathbb{R}^{H\times W}$ is an aggregated mask obtained by taking element-wise addition over $\mask_{1:T}$, and $\mathbf B_{\text{global}}\in\mathbb{R}^{H\times W\times L}$ is an aggregated bounding box tensor obtained by taking element-wise maximum over $\mathbf B_{1:T}$.

Finally, we additionally impose a reconstruction loss $\mathcal L_\text{rec}$ that encourages the predicted instance masks to be similar to the ground-truths.
We implement this idea using perceptual loss~\cite{Johnson:ECCV2016:PerceptualLoss,ChenQ17, Wang17Arxiv, Cha17Arxiv},
which measures the distance of real and fake images in the feature space of a pre-trained CNN by
\begin{align}
    \mathcal{L}_\text{rec} = \sum_l \big\lVert \Phi_l (G_\glob) - \Phi_l(M_\glob) \big\rVert,
    \label{eqn:mask_perceptual}
\end{align}
where $\Phi_l$ is the feature extracted from the $l$-th layer of a CNN.
We use the VGG-19 network~\cite{Vgg} pre-trained on ImageNet~\cite{Imagenet}
in our experiments.
Since our input to the pre-trained network is a binary mask, we replicate masks to channel dimension and use the converted mask to compute Eq.~\eqref{eqn:mask_perceptual}.
We found that using the perceptual loss significantly improves stability of GAN training and the quality of object shapes, as discussed in \cite{ChenQ17, Wang17Arxiv, Cha17Arxiv}.

Combining Eq.\eqref{eqn:shape_instance_loss}, \eqref{eqn:shape_global_loss} and \eqref{eqn:mask_perceptual},
the overall training objective for the shape generator becomes 
\vspace{-0.1cm}  
\begin{align}
     &\mathcal{L}_{\text{shape}} =  
      \lambda_i \mathcal{L}_{\text{inst}} +    
      \lambda_g \mathcal{L}_{\text{global}} +  
      \lambda_r \mathcal{L}_{\text{rec}},      
\label{eqn:obj_shape_generator}
\end{align}
where $\lambda_i, \lambda_g$ and $\lambda_r$ are hyper-parameters
that balance different losses,
which are set to 1, 1 and 10 in the experiment, respectively.
We provide more details of training and network architecture in the appendix (Section~\ref{sec:shape_detail}). 
\medskip   

\cutsectionup
\section{Synthesizing Images from Text and Layout}
\label{sec:pixel_gen}
\cutsectiondown

The outputs from the layout generator define location, size, shape and class information of objects, 
which provide semantic structure of a scene relevant to text.
Given the semantic structure and text, the objective of the image generator is to generate an image
that conforms to both conditions.
To this end, we first aggregate binary object masks $\mask_{1:T}$ to a semantic label map $\mathbf M \in \{0, 1\}^{H \times W \times L}$,
such that $\mathbf M_{ijk} = 1$ if and only if there exists an object of class $k$
whose mask $M_t$ covers the pixel $(i,j)$.
Then, given the semantic layout $\mathbf M$ and the text $\txt$,
the image generator is defined by
\begin{equation}
    \widehat{\img} = \Gimg(\mathbf M, \txt, \mathbf{z}),
    \label{eqn:image_generator}
\end{equation}
where $\mathbf z\sim\mathcal{N}(0,I)$ is a random noise.
In the following, we describe the network architecture
and training procedures of the image generator.

\paragraph{Model.}
Figure~\ref{fig:image_generator} illustrates the overall architecture of the image generator.
Our generator network is based on a convolutional encoder-decoder network~\cite{Isola17CVPR} with several modifications.
It first encodes the semantic layout $\mathbf M$ through several down-sampling layers
to construct a layout feature $\mathbf A\in\mathbb{R}^{h \times w\times d}$.
We consider that the layout feature encodes various context information of the input layout along the channel dimension.
To adaptively select a context relevant to the text, we apply attention on the layout feature.
Specifically, we compute a $d$-dimensional vector from the text embedding, and spatially replicate it to construct $\mathbf S\in\mathbb{R}^{h \times w\times d}$.
Then we apply gating on the layout feature
by $\mathbf A^g = \mathbf A \odot \sigma(\mathbf S)$, where $\sigma$ is the sigmoid nonlinearity, and $\odot$ denotes element-wise multiplication.
To further encode text information on background, we compute another text embedding with separate fully-connected layers and spatially replicate it to size $h\times w$.
The gated layout feature $\mathbf A^g$, the text embedding and noises are then combined by concatenation along channel dimension, and subsequently fed into several residual blocks and decoder to be mapped to an image.
We employ a cascaded network~\cite{ChenQ17} for decoder,
which takes the semantic layout $\mathbf M$ as an additional input to every upsampling layer.
We found that cascaded network enhances conditioning on layout structure and produces better object boundary.

For the 
discriminator network $D_{\text{img}}$, 
we first concatenate the generated image $X$ and the semantic layout $\mathbf M$. 
It is fed through a series of down-sampling blocks,
resulting in a feature map of size $h' \times w'$.
We concatenate it with a spatially tiled text embedding,
from which we compute a decision score of the discriminator.
\paragraph{Training.}
Conditioned on both the semantic layout $\mathbf M$ and
the text embedding $\txt$, the image generator $\Gimg$ is
jointly trained with the discriminator $\Dimg$. 
We define the objective function by 
$
\mathcal{L}_{\text{img}} = \lambda_a \mathcal{L}_{\text{adv}} + \lambda_r \mathcal{L}_{\text{rec}},
$
where
\begin{align}
%
%
%
    %
    \mathcal L_{\text{adv}} &=
        \mathbb E_{ (\mathbf M, \txt, \img) } \Big[
            \log D_{\text{img}} \big( \mathbf M, \txt, \img \big)
        \Big]
    \label{eqn:image_adv_loss} \\
    &~~~ +
        \mathbb E_{(\mathbf M, \txt), \mathbf{z} } \Big[
            \log \Big( 1 - D_{\text{img}} \big(\mathbf M, \txt, \Gimg(\mathbf M, \txt, \mathbf{z}) \big) \Big)
        \Big]
    , \nonumber \\
    \mathcal{L}_{\text{rec}} &=  \sum_l \lVert \Phi_l (\Gimg(\mathbf M, \txt, \mathbf{z})) - \Phi_l(\img) \rVert,
\label{eqn:obj_pixel_generator}
\vspace{-0.1cm}
\end{align}
where $\img$ is a ground-truth image associated with semantic layout $\mathbf M$.
As in the mask generator, we apply the same perceptual loss $\mathcal L_\text{rec}$,
which is found to be effective.
We set the hyper-parameters $\lambda_a=1$, $\lambda_r=10$ in our experiment.
More details on network architecture and training procedure is provided in appendix (Section~\ref{sec:image_detail}).
%

%

%

%
%
%

%
%


\begin{table*}[t]
\vspace{-0.2cm}
\centering
\setlength\doublerulesep{0.5pt}
\small
\begin{tabular}{l|cc|cccccc|c}
&                                         &
& \multicolumn{6}{c|}{Caption generation}   &
\multirow{2}{*}{\makecell{Inception \\ \cite{Salimans:NIPS2016:ImprovedGan}}} \\
\multicolumn{1}{c|}{Method}                & Box   & Mask  & BLEU-1         & BLEU-2         & BLEU-3         & BLEU-4         & METEOR         & CIDEr          \\
\hline \hline
Reed \etal \cite{ReedS16ICML}              & -     & -     & 0.470          & 0.253          & 0.136          & 0.077          & 0.122          & 0.160           & 7.88 $\pm$ 0.07           \\
StackGAN   \cite{HanZ17}                   & -     & -     & 0.492          & 0.272          & 0.152          & 0.089          & 0.128          & 0.195           & 8.45 $\pm$ 0.03           \\
\hline
Ours   & Pred. & Pred. & \textbf{0.541} & \textbf{0.332} & \textbf{0.199} & \textbf{0.122} & \textbf{0.154} & \textbf{0.367}  & \textbf{11.46 $\pm$ 0.09} \\
\hline \hline
\multirow{2}{*}{Ours (control experiment)} & GT    & Pred. & 0.556          & 0.353          & 0.219          & 0.139          & 0.162          & 0.400           & 11.94 $\pm$ 0.09          \\
& GT    & GT    & 0.573          & 0.373          & 0.239          & 0.156          & 0.169          & 0.440           & 12.40 $\pm$ 0.08          \\
\hline
Real images (upper bound)                  & -     & -     & 0.678          & 0.496          & 0.349          & 0.243          & 0.228          & 0.802           & -                         \\
\hline
\end{tabular}
\medskip
\caption{
    Quantitative evaluation results.
    Two evaluation metrics based on caption generation and
    the Inception score are presented.
    The second and third columns indicate types of bounding box or mask layout used in image generation, where ``GT'' indicates ground-truth and ``Pred.'' indicates predicted one by our model.
    The last row presents the caption generation performance 
    on real images, which corresponds to upper-bound of caption generation metric.
    Higher is better in all columns. 
}
\label{tab:eval_caption}
\vspace{-0.1cm}
\end{table*}

\section{Experiments}
\label{sec:experiment}
\cutsectiondown

\ifdefined\paratitle {\color{blue}
[Dataset: mscoco] \\
} \fi

\subsection{Experimental Setup}
\label{sec:implementation}
\cutsubsectiondown

\cutparagraphup
\paragraph{Dataset.}
We use the MS-COCO dataset~\cite{Lin:2014:MSCOCO} to evaluate our model.
It contains 164,000 training images over 80 semantic classes, where each image is associated with instance-wise annotations (\ie, object bounding boxes and segmentation masks) and 5 text descriptions.
The dataset has complex scenes with many objects in a diverse context, which makes generation very challenging.
We use the official train and validation splits from MS-COCO 2014 for training and evaluating our model, respectively.

\begin{table}[!t]
    \centering
    \small
    \begin{tabular}{l|c|c}
        \multicolumn{1}{c|}{Method}   & \makecell{ratio of \\ [-0.5ex]ranking 1st}   & \makecell{\emph{vs.} Ours} \\ 
        \hline
        StackGAN~\cite{HanZ17}        & 18.4 \%                                      & 29.5 \% \\
        Reed \etal~\cite{ReedS16ICML} & 23.3 \%                                      & 32.3 \% \\
        Ours                          & \textbf{58.3} \%                             & - \\
    \end{tabular}
    
    \smallskip
    \caption{
    Human evaluation results.
    }
    \label{tab:human_eval}
    \vspace{-0.6cm}
\end{table}

\begin{figure*}[htb] \begin{center}
    \vspace{-0.2cm}
    \includegraphics[width=\textwidth]{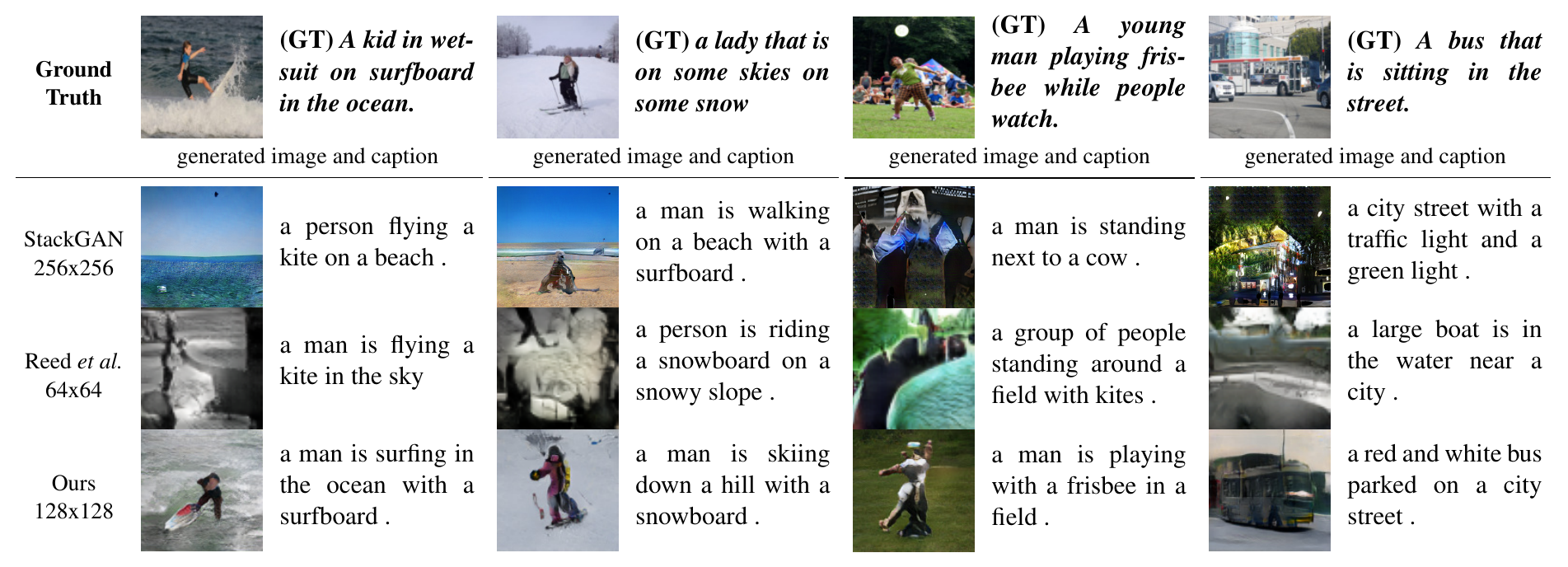}
    \vspace{-0.5cm}
    \caption{
        Qualitative examples of generated images conditioned on text descriptions on the MS-COCO validation set,
        using our method and baselines (StackGAN \cite{HanZ17} and Reed \etal~\cite{ReedS16ICML}).
        The input text and ground-truth image are shown in the first row.
        For each method, we provide a reconstructed caption conditioned on the generated image.
    }
\label{fig:caption_baselines}
\vspace{-0.8cm}
\end{center} \end{figure*}

\cutparagraphup
\paragraph{Evaluation metrics.}
We evaluate text-conditional image generation performance using various metrics:
Inception score, 
caption generation, and human evaluation.

\textit{Inception score ---}
We compute the Inception score~\cite{Salimans:NIPS2016:ImprovedGan} by applying pre-trained classifier on synthesized images and investigating statistics of their score distributions.
It measures recognizability and diversity of generated images, and has been known to be correlated with human perceptions on visual quality
\cite{Odena17ICML}.
We use the Inception-v3~\cite{Szegedy16CVPR} network pre-trained on ImageNet \cite{Imagenet} for evaluation, and measure the score for all validation images.

\textit{Caption generation ---}
In addition to the Inception score,
assessing performance of text-conditional image generation necessitates measuring the relevance of generated image to the input text.
To this end, we generate sentences from the synthesized image and measure the similarity between input text and predicted sentence.
The underlying intuition is that if the generated image is relevant to input text and its contents are recognizable,
one should be able to guess the original text from the synthesized image.
We employ an image caption generator~\cite{Vinyals15CVPR} trained on MS-COCO to generate sentences,
where one sentence is generated per image by greedy decoding. 
We report three standard 
language similarity metrics:
BLEU~\cite{Papineni:ACL2002:BLEU}, METEOR~\cite{Banerjee:ACL2005:METEOR} and CIDEr~\cite{Vedantam:CVPR2015:CIDEr}.
\textit{Human evaluation ---}
Evaluation based on caption generation is beneficial for large-scale evaluation but may introduce unintended bias by the caption generator.
To verify the effectiveness of caption-based evaluation, we conduct human evaluation using Amazon Mechanical Turk.
For each text randomly selected from MS-COCO validation set, we presented 5 images generated by different methods, and asked users to rank the methods based on the relevance of generated images to text.
We collected results for 1000 sentences, each of which is annotated by 5 users.
We report results based on the ratio of each method ranked as the best, and one-to-one comparison between ours and the baselines.

\subsection{Quantitative Analysis}
\label{sec:comparison}
\cutsubsectiondown

We compare our method with two state-of-the-art approaches~\cite{ReedS16ICML,HanZ17} based on conditional GANs. 
Table~\ref{tab:eval_caption} and Table~\ref{tab:human_eval} summarizes the quantitative evaluation results.

\begin{figure*}[t] \begin{center}
    \includegraphics[width=\textwidth]{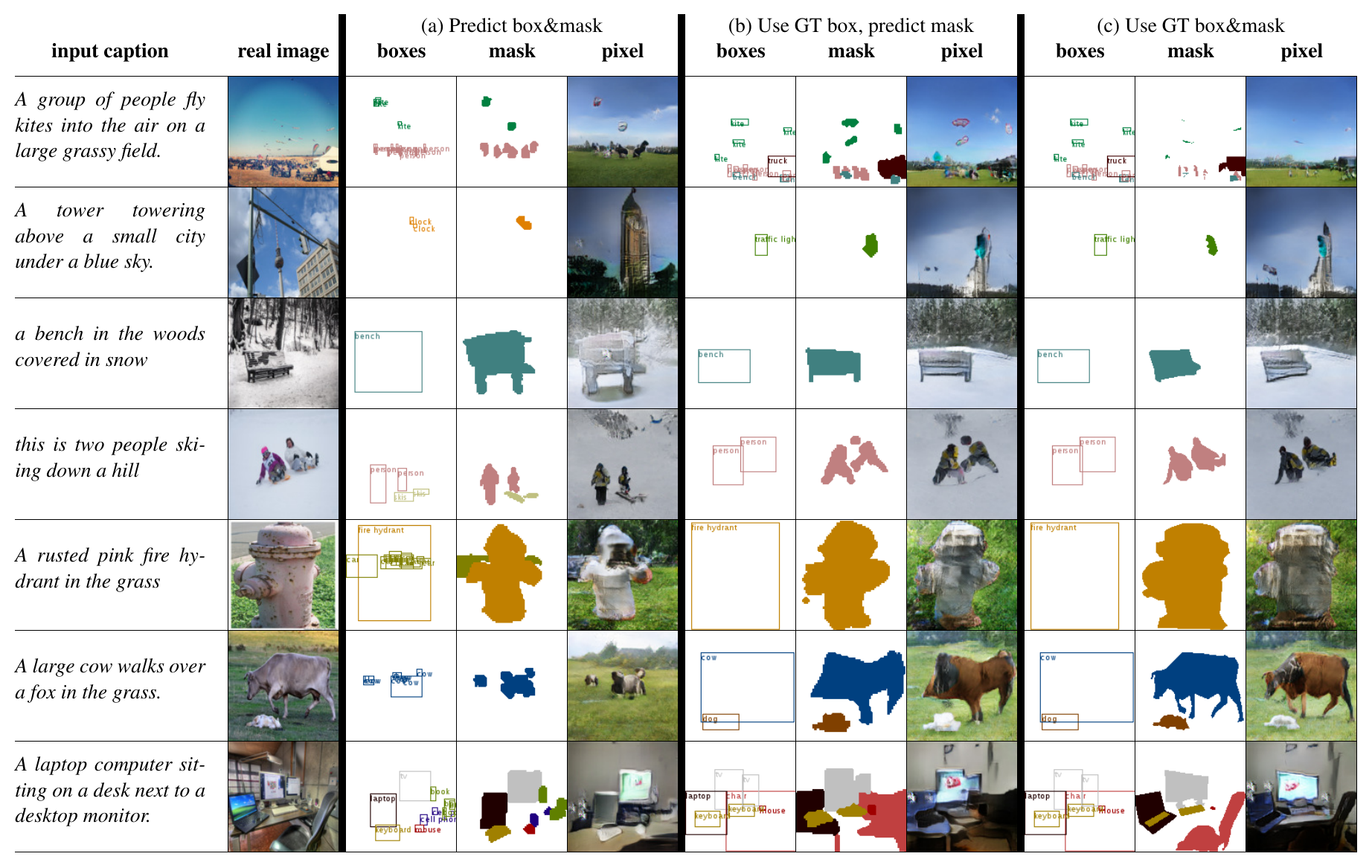}
    
    \vspace{-0.15cm}
    \caption{
    Image generation results of our method. Each column corresponds to generation results conditioned on (a) predicted box and mask layout, (b) ground-truth box and predicted mask layout and (c) ground-truth box and mask layout.
    Classes are color-coded for illustration purpose.
    See Figure~\ref{fig:samples_box_mask_image_supp} for more examples on the generated layouts and images.
    Best viewed in color.
}
\label{fig:samples_component_study}
\vspace{-0.6cm}
\end{center} \end{figure*}

\paragraph{Comparisons to other methods.}
We first present systemic evaluation results based on Inception score and caption generation performance. The results are summarized in Table~\ref{tab:eval_caption}.
The proposed method substantially outperforms existing approaches based on both evaluation metrics.
In terms of Inception score, our method outperforms the existing approaches with a substantial margin, presumably because our method generates  more recognizable objects.
Caption generation performance shows that captions generated from our synthesized images
 are more strongly correlated with the input text than the baselines.
This shows that images generated by our method are better aligned with descriptions and are easier to recognize semantic contents.

Table~\ref{tab:human_eval} 
summarizes comparison results based on human evaluation.
When users are asked to rank images based on their relevance to input text,
they choose images generated by our method as the best in about $60\%$ of all presented sentences, which is substantially higher than baselines (about $20\%$).
This is consistent with the caption generation results in Table~\ref{tab:eval_caption}, in which our method substantially outperforms the baselines while their performances are comparable.

%
\begin{figure}[t]
\centering
\centering
\includegraphics[width=0.95\linewidth]{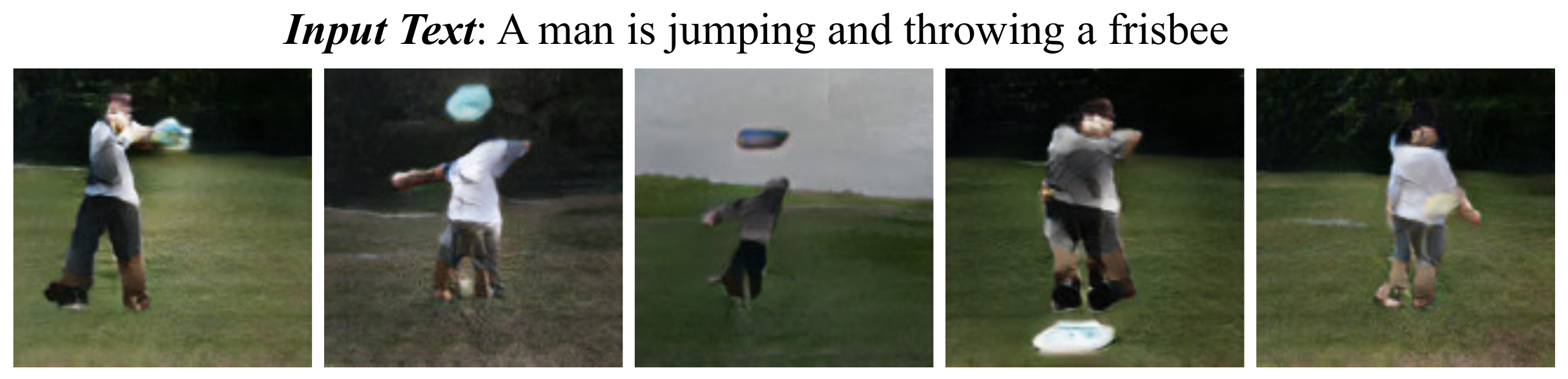} \\
\includegraphics[width=0.95\linewidth]{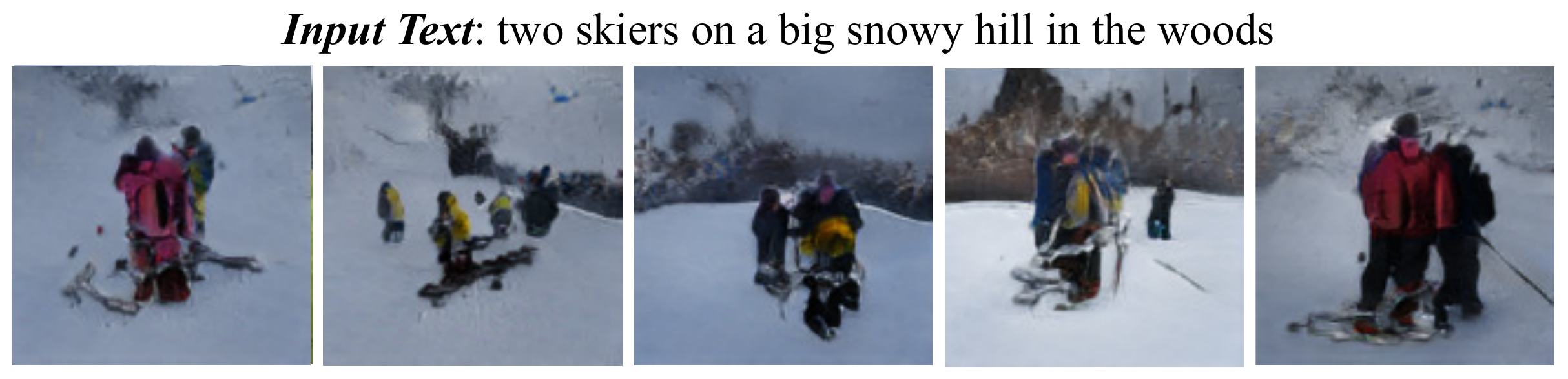} \\
\includegraphics[width=0.95\linewidth]{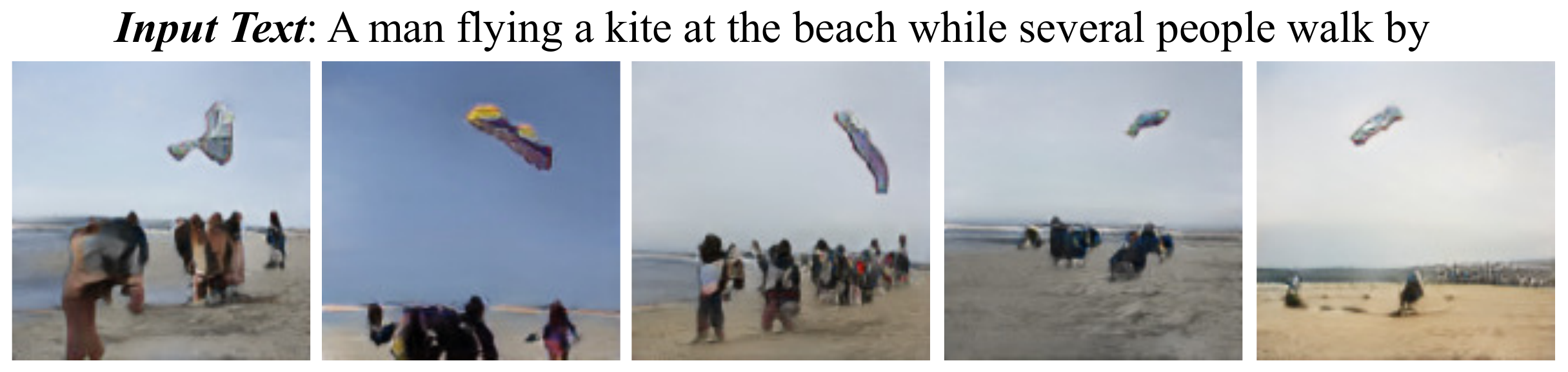}

\caption{\small Multiple samples generated from a text description. See Figure~\ref{fig:diversity_supp} for more results.}

\vspace{-0.4cm}
\label{fig:diversity}
\end{figure}

Figure~\ref{fig:caption_baselines} illustrates qualitative comparisons.
Due to adversarial training, images generated by the other methods, especially StackGAN \cite{HanZ17},
tend to be clear and exhibits high frequency details.
However, it is difficult to recognize contents from the images,
since they often fail to predict important semantic structure of object and scene.
As a result, the reconstructed captions from the generated images are usually not relevant to the input text.
Compared to them,
our method generates much more recognizable and semantically meaningful images by conditioning the generation with inferred semantic layout,
and is able to reconstruct descriptions that better align with the input sentences.

%

\begin{figure}[!t]
\vspace{-0.1cm}
\centering
\includegraphics[width=1\linewidth]{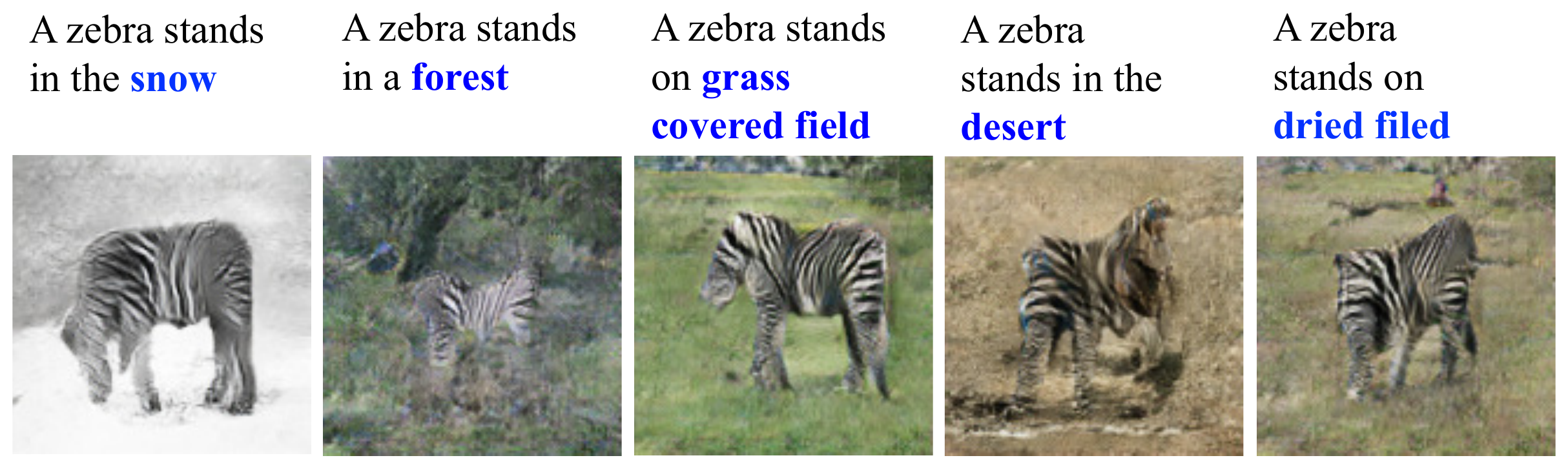} \\
\includegraphics[width=1\linewidth]{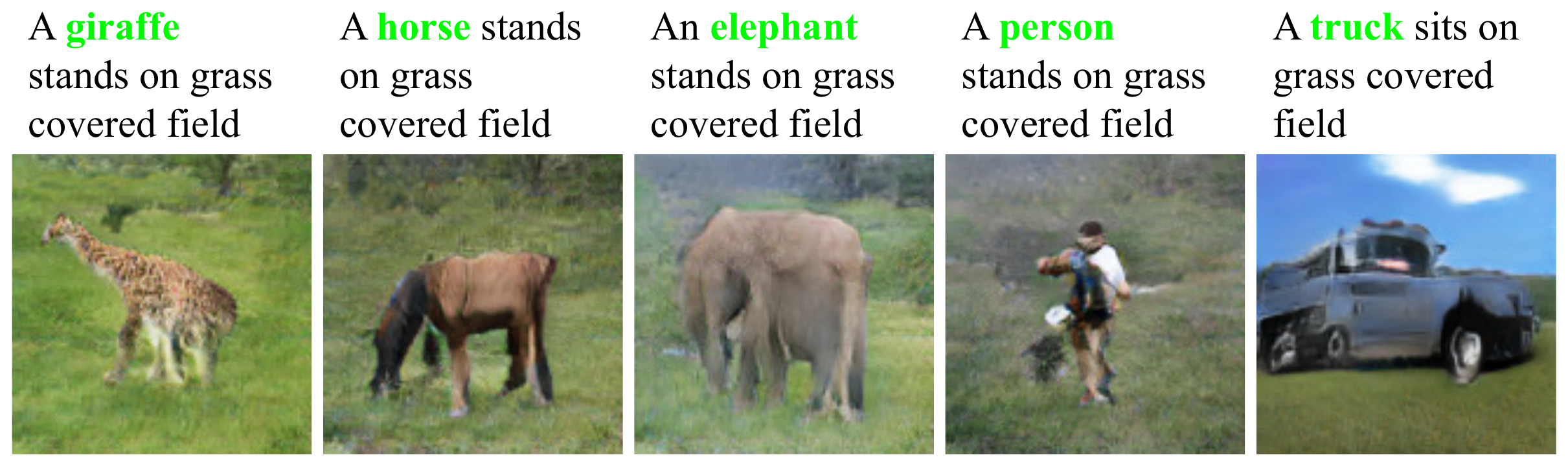} 
\includegraphics[width=1\linewidth]{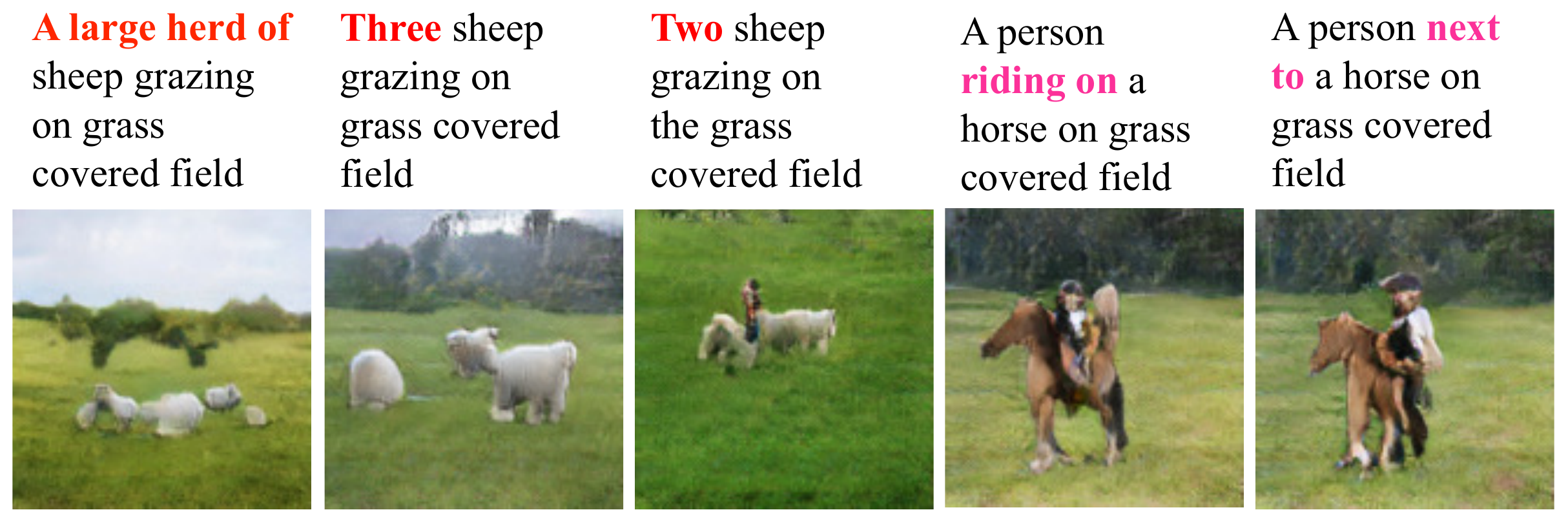} 
\caption{Generation results by manipulating captions.
The manipulated parts of texts are highlighted in \textbf{bold} characters, where the types of manipulation is indicated by different colors.
{\bf \color{blue} Blue}: scene context, {\bf \color{magenta} Magenta}: spatial location, {\bf \color{red} Red}: the number of objects,  {\bf \color{green} Green}: object category.
}
\vspace{-0.3cm}
\label{fig:text_conditional}
\end{figure}

%
%
\cutparagraphup
\paragraph{Ablative Analysis.}
To understand quality and the impact of the predicted semantic layout, we conduct an ablation study by gradually replacing the bounding box and mask layout predicted by layout generator with the ground-truths.
Table~\ref{tab:eval_caption} summarizes quantitative evaluation results.
As it shows, replacing the predicted layouts to ground-truths leads with gradual performance improvements, which shows predictions errors in both bounding box and mask layout.

\ifdefined\paratitle {\color{blue}
[Table1: caption reconstruction. Evaluation metric, compared methods] \\
} \fi

\ifdefined\paratitle {\color{blue}
[Table2: inception score. entropy and cross-entropy.] \\
} \fi

\subsection{Qualitative Analysis}
\label{sec:qualitative}
\cutsubsectiondown

Figure~\ref{fig:samples_component_study} 
shows qualitative results of our method.
For each text, we present the generated images alongside the predicted semantic layouts.
As in the previous section, we also present our results conditioned on ground-truth layouts.
As it shows, our method generates reasonable semantic layout and image matching the input text;
it generates bounding boxes corresponding to fine-grained scene structure implied in texts (\ie object categories, the number of objects),
and object masks capturing class-specific visual attributes as well as relation to other objects.
Given the inferred layouts, our image generator produces correct object appearances and background compatible with text.
Replacing the predicted layouts with ground-truths makes the generated images to have a similar context to original images.
%

\begin{figure}[!t]
\vspace{-0.1cm}
\centering
\small
\includegraphics[width=1\linewidth]{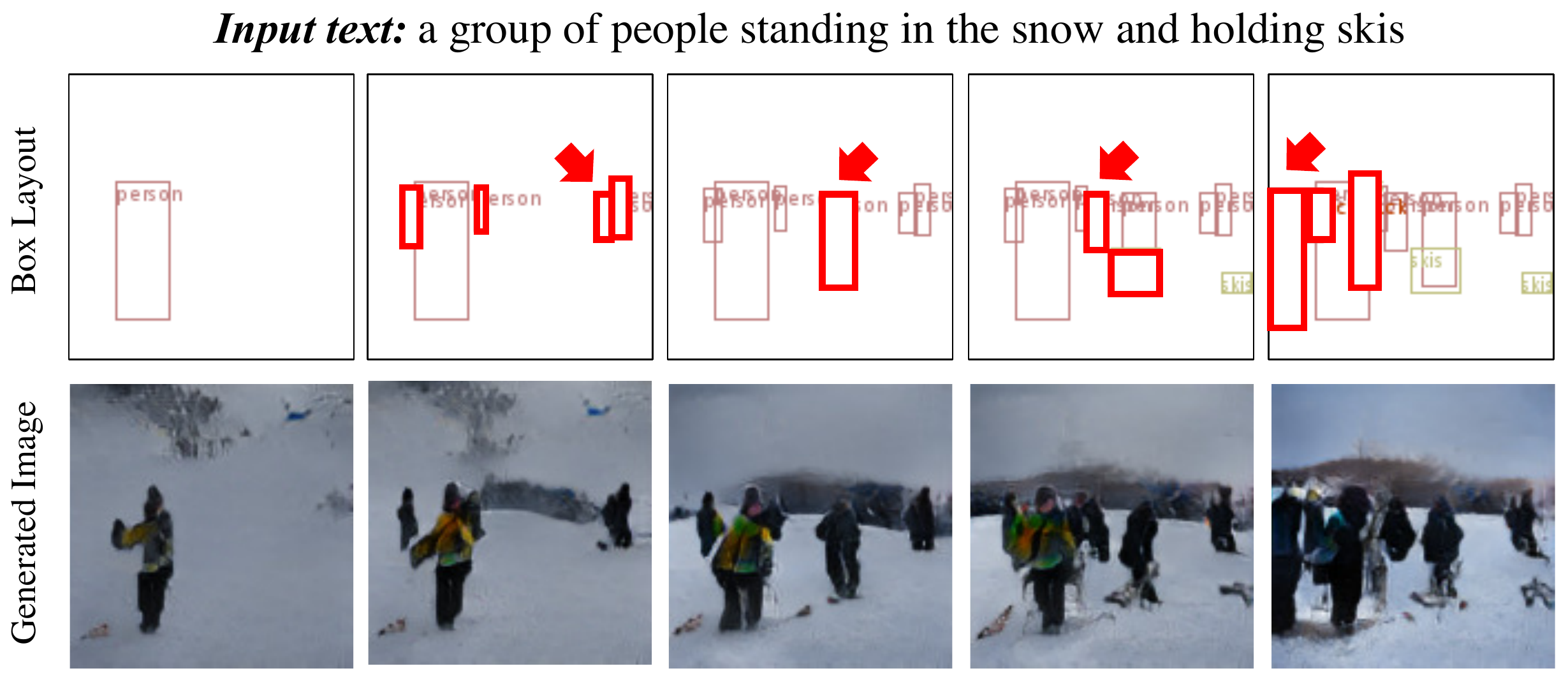} \\
(a) Generation results by adding new objects.
\\[0.7ex]
\includegraphics[width=1\linewidth]{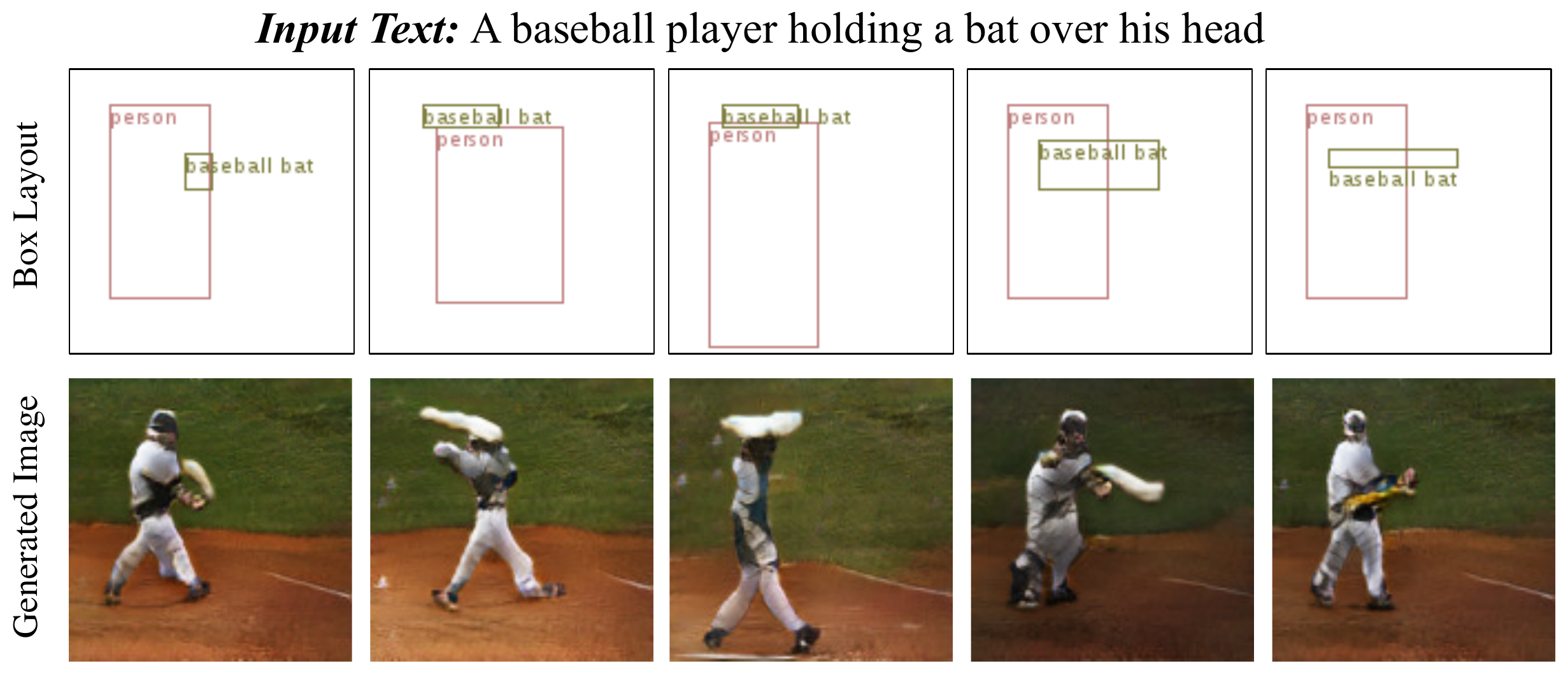} 
\mbox{(b) Generation results by changing spatial configuration of objects.}\\

\caption{
Examples of controllable image generation.
See Figure~\ref{fig:controllable_gen_add_supp} and \ref{fig:controllable_gen_move_supp} for more results.
}
\label{fig:controllable_gen}
\vspace{-0.3cm}
\end{figure}

\cutparagraphup
\paragraph{Diversity of samples.}
To assess the diversity in generation, we sample multiple images while fixing the input text.
Figure~\ref{fig:diversity} illustrates the example images generated by our method.
Our method generates diverse semantic structures given the same text description, while preserving semantic details such as the number of objects and object categories.
%


\cutparagraphup
\paragraph{Text-conditional generation.}
To see how our model incorporates text description in generation process, we generate images while modifying parts of the descriptions.
Figure~\ref{fig:text_conditional} illustrates the example results.
When we change the context of descriptions such as object class, number of objects, spatial composition of objects and background patterns, our method correctly adapts semantic structure and images based on the modified part of the text.

\cutparagraphup
\paragraph{Controllable image generation.}
We demonstrate controllable image generation by modifying bounding box layout.
Figure~\ref{fig:controllable_gen} illustrates the example results. 
Our method updates object shapes and context based on the modified semantic layout (\eg adding new objects, changing spatial configuration of objects) and generates reasonable images. 
See Figure~\ref{fig:controllable_gen_add_supp} and \ref{fig:controllable_gen_move_supp} for more examples on various types of layout modifications.

\smallskip
\cutsectionup
\section{Conclusion}
\label{sec:conclusion}
\cutsectiondown

We proposed an approach for text-to-image synthesis which explicitly infers and exploits a semantic layout as an intermediate representation from text to image.
Our model hierarchically constructs a semantic layout in a coarse-to-fine manner by a series of generators. 
By conditioning image generation on explicit layout prediction,
our method generates complicated images that preserve semantic details and highly relevant to the text description. 
We also showed that the predicted layout can be used to control generation process.
We believe that end-to-end training of layout and image generation would be an interesting future work.
%


\cutparagraphup
\paragraph{Acknowledgments}
This work was supported in part by ONR N00014-13-1-0762, NSF CAREER IIS-1453651, DARPA Explainable AI (XAI) program \#313498, and Sloan Research Fellowship.

{\small
\bibliographystyle{ieee}
\bibliography{composit_gen.bib}
}

\clearpage
\appendix
\section*{Appendix}
\addcontentsline{toc}{section}{Appendix}
\renewcommand{\thesection}{\Alph{section}}

\section{Implementation Details}
\label{sec:details}

\subsection{Box Generator}
\label{sec:box_detail}
This section describes the details of the box generator.
Denoting bounding box of $t$-th object as $B_t=(b^x_{t}, b^y_{t}, b^w_{t}, b^h_{t}, \cls_t)$, the joint probability of sampling $B_t$ from the box generator is given by
\begin{align}
p(b^x_{t}, b^y_{t}, b^w_{t}, b^h_{t}, \cls_t) &= p(\cls_t)p(b^x_{t}, b^y_{t}, b^w_{t}, b^h_{t}| \cls_t). 
\label{eqn:box_generator_obj_supp}
\end{align}
We drop the conditioning variables for notational brevity.
As described in the main paper, we implement $p(\cls_t)$ by categorical distribution and $p(b^x_{t}, b^y_{t}, b^w_{t}, b^h_{t} | \cls_t)$ by a mixture of quadravariate Gaussians.
However, modeling full convariance matrix of quadravariate Gaussian is expensive as it involves many parameters. 
Therefore, we decompose the box coordinate probability as $p(b^x_{t}, b^y_{t}, b^w_{t}, b^h_{t}| \cls_t) = p(b^x_{t}, b^y_{t}|\cls_t)p(b^x_{t}, b^y_{t}| b^w_{t}, b^h_{t}, \cls_t)$, and approximate it with two bivariate Gaussian mixtures by
\begin{align}
p(b^x_{t}, b^y_{t} | \cls_t) &= \sum_{k=1}^K \boldsymbol \pi^{xy}_{t,k} \mathcal{N} \left( b^x_{t}, b^y_{t} ; \boldsymbol \mu^{xy}_{t,k}, \boldsymbol \Sigma^{xy}_{t,k} \right), \nonumber \\
p(b^w_{t}, b^h_{t} | b^x_{t}, b^y_{t}, \cls_t) &= \sum_{i=k}^K  \boldsymbol \pi^{wh}_{t,k} \mathcal{N} \left( b^w_{t}, b^h_{t} ; \boldsymbol \mu^{wh}_{t,k}, \boldsymbol \Sigma^{wh}_{t,k} \right) \nonumber.
\end{align}
Then the parameters for Eq.~\eqref{eqn:box_generator_obj_supp} are obtained from LSTM outputs at each step by
\begin{align}
\left[ h_t, c_t \right] &= \text{LSTM}(B_{t-1};\left[h_{t-1}, c_{t-1}\right]), \\
\cls_t &= W^lh_t + \mathbf b^l, \\
\boldsymbol \theta^{xy}_{t} &= W^{xy}[h_t, \cls_t] + \mathbf b^{xy}, \\
\boldsymbol \theta^{wh}_{t} &= W^{wh}[h_t, \cls_t, b_x, b_y] + \mathbf b^{wh},
\end{align}
where $\boldsymbol \theta^{\boldsymbol \cdot}_{t}=[\boldsymbol \pi^{\boldsymbol \cdot}_{t,1:K}, \boldsymbol \mu^{\boldsymbol \cdot}_{t,1:K}, \boldsymbol \Sigma^{\boldsymbol \cdot}_{t,1:K}]$ are the parameters for GMM concatenated to a vector. 

For training, we employ an Adam optimizer~\cite{kingma2014adam} with learning rate 0.001, $\beta_1 = 0.9, \beta_2 = 0.999$ and exponentially decrease the learning rate with rate 0.5 at every epoch after the initial 10 epochs.

\subsection{Shape Generator}
\label{sec:shape_detail}
\begin{figure*}[!t]
    \centering
    \includegraphics[width=1\textwidth]{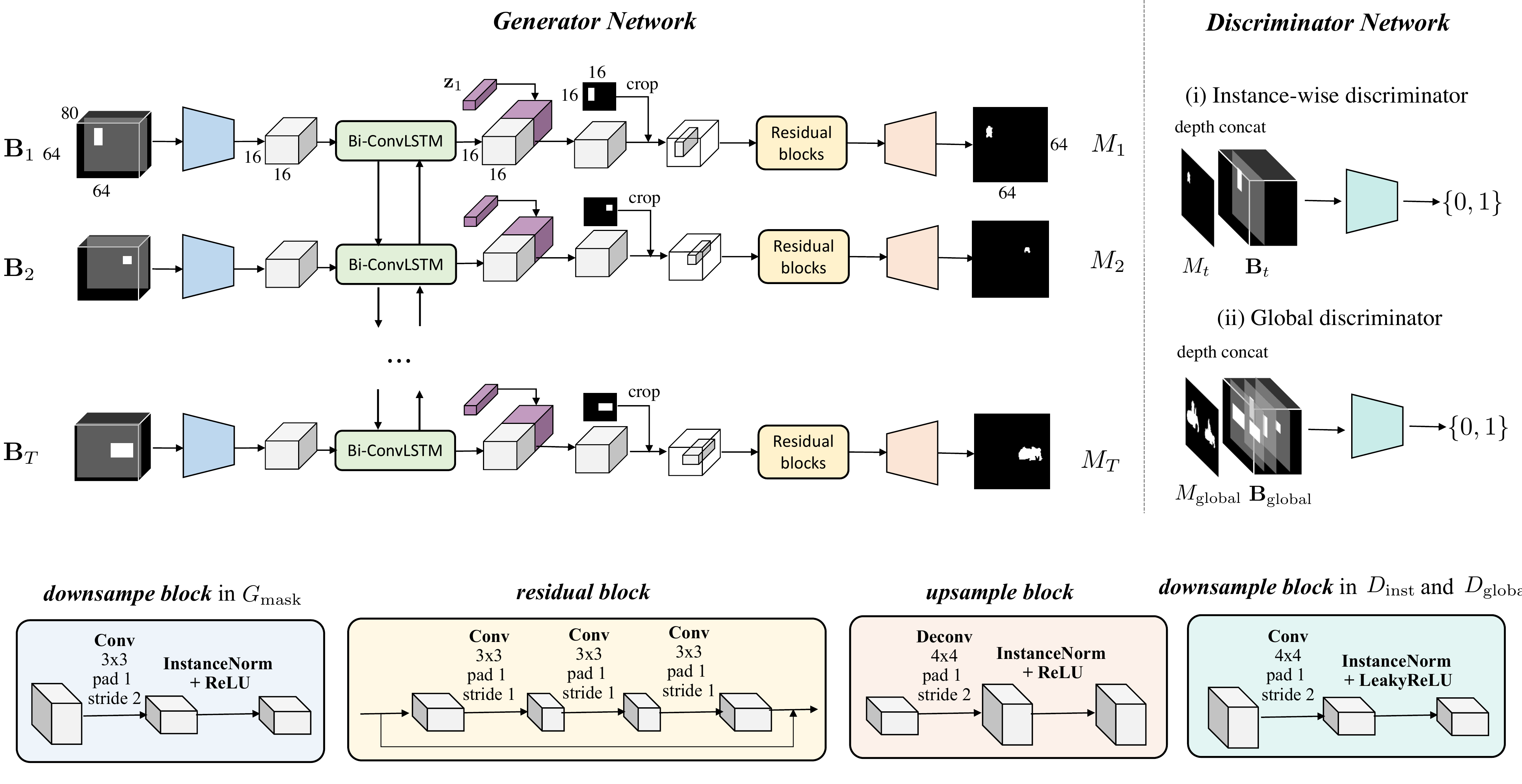}
    \caption{Architecture of the shape generator.}
    \label{fig:shape_generator_detail}
    \vspace{3.0cm}
\end{figure*}

\begin{figure*}[!t]
    \centering
    \includegraphics[width=1\textwidth]{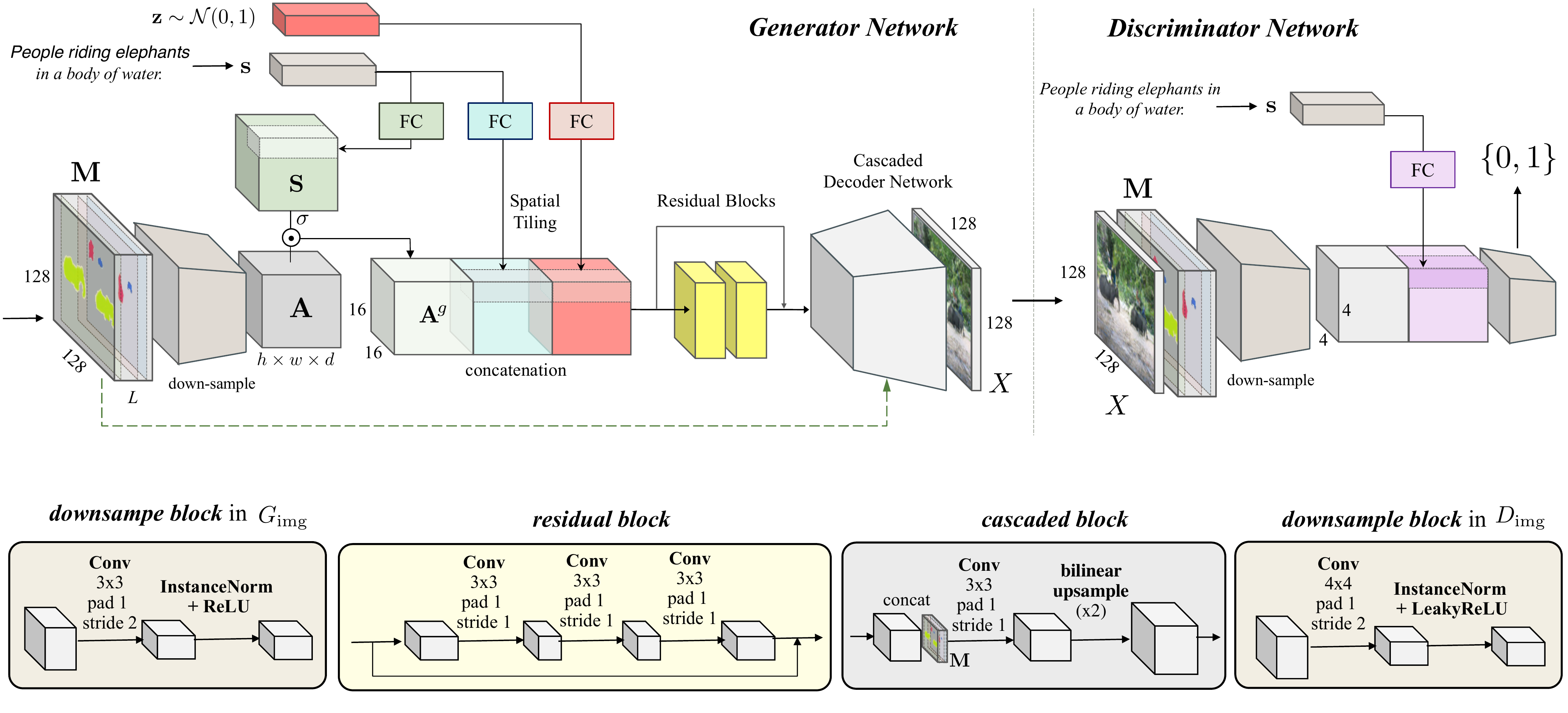}
    \caption{Architecture of the image generator.}
    \label{fig:image_generator_detail}
\end{figure*}

We provide a detailed architecture of the shape generator $\Gmask$ and the two discriminators $D_\inst$ and $D_\glob$ in Figure~\ref{fig:shape_generator_detail}.
At each step $t$, we encode a box tensor $\mathbf B_t$ by a series of downsampling layers, where each downsampling layer is implemented by a stride-2 convolution followed by instance-wise normalization~\cite{instnorm} and ReLU.
The encoded feature is fed into the bidirectional convolutional LSTM (bi-convLSTM), and combined with features from all object instances.
On top of the bi-convLSTM output at each step $t$, we add a noise $\mathbf z_t$ by spatial replication and depth concatenation, and apply masking operation so that regions outside the object bounding box $B_t$ are all set to 0.
The masked feature is fed into several residual blocks, and mapped to a binary mask $\mask_t$ by a series of upsampling layers.
Similar to downsampling layers, we implement an upsampling layer by stride-2 deconvolution followed by instance-wise normalization and ReLU except the last one, which is $1\times 1$ convolution followed by the sigmoid nonlinearity.

The instance-wise discriminator $D_\inst$ and global discriminator $D_\glob$ share the same architecture but have separate parameters. 
The input to the instance-wise discriminator is constructed by concatenating the box tensor $\mathbf B_t$ and the corresponding binary mask $\mask_t$ through channel dimension, while the one for global discriminator is constructed by concatenating the aggregated box tensor $\mathbf B_\glob$ and the aggregated masks $\mask_\glob$. 
Both discriminators encode the input by a series of downsampling layers, which are implemented by stride-2 convolutions followed by instance-wise normalization and Leaky-ReLU~\cite{leakyrelu}.

For training, we employ an Adam optimizer~\cite{kingma2014adam} with learning rate 0.0002, $\beta_1=0.5, \beta_2=0.999$ and linearly decrease the learning rate after the first 50-epochs training.

\subsection{Image Generator}
\label{sec:image_detail}
A detailed architecture of the image generator is illustrated in Figure~\ref{fig:image_generator_detail}.
The architecture of the downsampling and the residual blocks are same as the ones used in the shape generator. 
To encourage the model to generate images that match the input layout, we implement upsampling layers based on cascaded refinement network~\cite{ChenQ17}.
At each upsampling layer, it takes an output from the previous layer and the semantic layout resized to the same spatial size as inputs, and combines them by depth concatenation followed by convolution.
The combined feature map is then spatially upscaled by bilinear upsampling followed by instance-wise normalization and ReLU, and subsequently fed into the next upsampling layer. 

To encourage the model to generate images that match input text descriptions, we employ a matching-aware loss proposed in \cite{ReedS16ICML}.
Denoting a ground-truth training example as $(\mathbf M, \txt, \img)$, where $\mathbf M$, $\txt$ and $\img$ denote semantic layout, text embedding and image, respectively, we construct an additional mismatching triple $(\mathbf M, \widetilde{\txt}, \img)$ by sampling random text embedding $\widetilde{\txt}$ non-relevant to the image.
We consider it as additional fake examples in adversarial training, and extend the conditional adversarial loss for image generator (Eq.~\eqref{eqn:image_adv_loss} in the main paper) as
\begin{align}
    \mathcal L_{\text{adv}} &=
        \mathbb E_{ (\mathbf M, \txt, \img) } \Big[
            \log D_{\text{img}} \big( \mathbf M, \txt, \img \big)
        \Big]
    \label{eqn:image_adv_loss_supp} \\
        &~~~ + \mathbb E_{ (\mathbf M, \widetilde{\txt}, \img) } \Big[
            \log \Big( 1 - D_{\text{img}} \big( \mathbf M, \widetilde{\txt}, \img \big) \Big)
        \Big] \nonumber    \\
    &~~~ +
        \mathbb E_{(\mathbf M, \txt), \mathbf{z} } \Big[
            \log \Big( 1 - D_{\text{img}} \big(\mathbf M, \txt, \Gimg(\mathbf M, \txt, \mathbf{z}) \big) \Big)
        \Big]
    . \nonumber 
\label{eqn:obj_pixel_generator_detail}
\end{align}
We found that employing matching-aware loss substantially improves text-conditional generation and stabilizes overall GAN training.   

For training, we employ an Adam optimizer~\cite{kingma2014adam} with learning rate 0.0002, $\beta_1=0.5, \beta_2=0.999$ and linearly decrease the learning rate after the first 30-epoch training. 
%


\begin{table*}[!th] 
\centering
\bigskip 
\small
\begin{tabular}{l|cccccc}
& \multicolumn{6}{c}{Caption generation} \\
\multicolumn{1}{c|}{Method}                & BLEU-1         & BLEU-2         & BLEU-3         & BLEU-4         & METEOR         & CIDEr          \\
\hline
Ours (Full)   & \textbf{0.541} & \textbf{0.332} & \textbf{0.199} & \textbf{0.122} & \textbf{0.154} & \textbf{0.367}  \\
\hline \hline
w/o shape generator &0.494 &0.291 &0.170 &0.102 &0.131 &0.215 \\
w/o perceptual loss in shape generator  &0.536 &0.329 &0.196 &0.119 &0.151 &0.324\\
w/o perceptual loss in image generator  &0.491 &0.272 &0.146 &0.080 &0.126 &0.170   \\
w/o attention in image generator  &0.522 &0.315 &0.184 &0.112 &0.148 &0.324\\
\hline
\end{tabular}
\medskip
\caption{
    Ablation study of the proposed method.
    The first row corresponds to the performance of our model presented in the main paper.
}
\label{tab:ablation_study}
\bigskip
\bigskip
\end{table*}

\begin{figure*}[!th] 
\begin{center}
    \includegraphics[width=\textwidth]{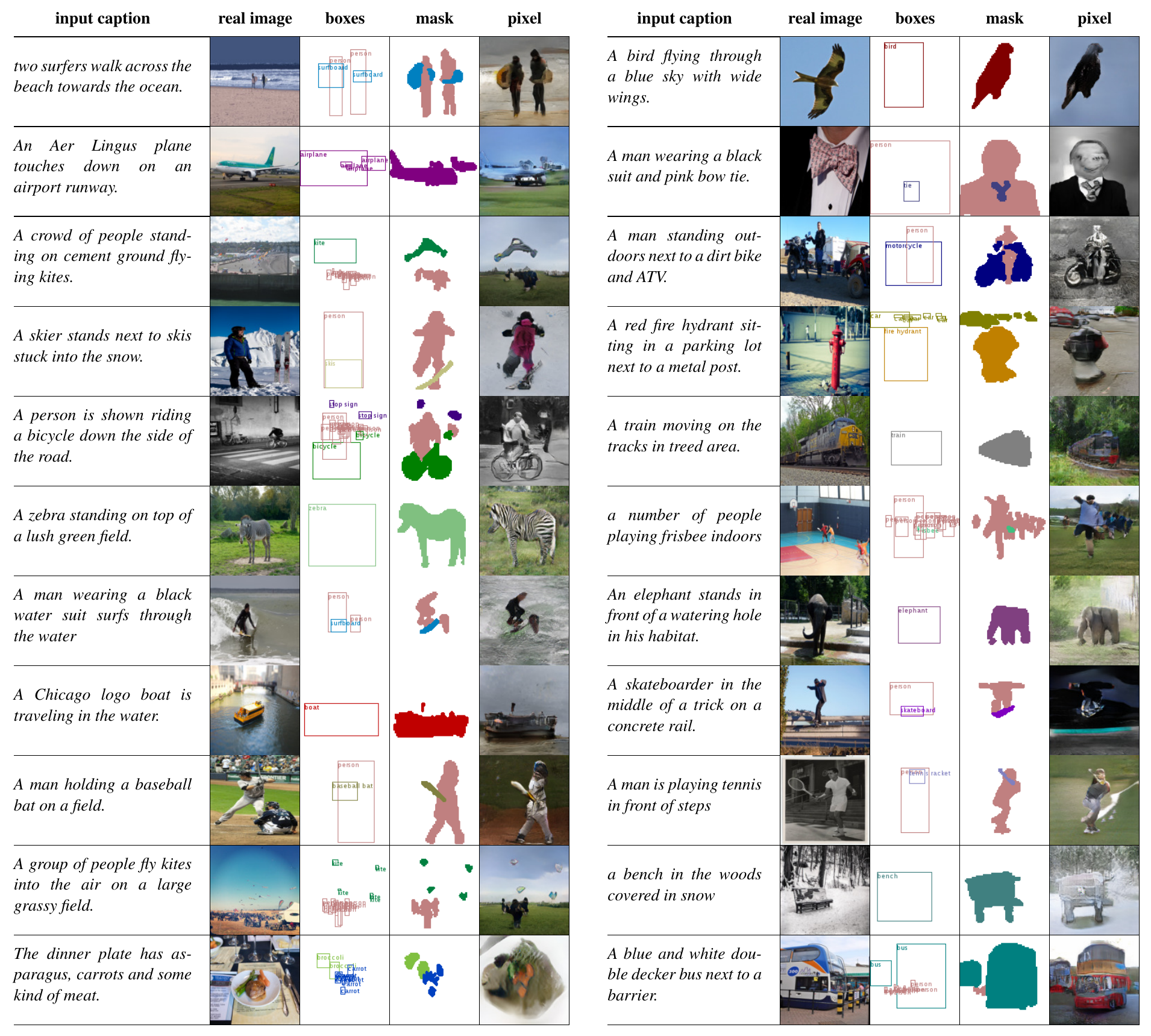}
    \caption{
    Illustrations of end-to-end prediction results of our method.
    Best viewed in color.
}
\label{fig:samples_box_mask_image_supp}
\end{center}
\end{figure*}
%
\medskip
\section{Additional Experiment Results}
\subsection{Ablative Analysis}
To understand the impact of each component in the proposed framework, we conduct an ablation study by varying configurations of the proposed model.
Table~\ref{tab:ablation_study} summarizes the results based on caption generation performance.

\paragraph{Impact of shape generator}
We first investigate the impact of shape generator. 
To this end, we remove the shape generator from our generation pipeline, and modify the image generator to generate images directly from box generator outputs.
Specifically, we feed the aggregated bounding box tensor $\mathbf B_\glob$ as an input to the image generator, which is constructed by taking pixel-wise maximum over all box tensors as $\mathbf B_\glob (i,j,l) = \max_t \mathbf B_t(i,j,l)$\footnote{Note that the the aggregated box tensor $\mathbf B_\glob$ can be considered as a semantic layout $\mathbf M$ that the shape of each object is a rectangular box.}. 
The result is presented in the second row in Table~\ref{tab:ablation_study}.
Removing the shape generator leads to substantial performance degradation, since predicting accurate object shapes and textures directly from bounding box is a complicated task; the image generator tends to miss detailed object shapes such as body parts, which are critical to recognize the image content for human.
By explicitly inferring object shapes, it improves the overall image quality and interpretability of content.

\paragraph{Impact of perceptual loss in shape generator}
To see the effectiveness of perceptual loss in shape generator, we train the model after replacing the reconstruction loss in Eq.~\eqref{eqn:mask_perceptual} with the $\ell_1$ loss 
on pixels.
The result is presented in the third row in Table~\ref{tab:ablation_study}.
As it shows, adding perceptual loss to the shape generator improves the accuracy of object shapes and leads to more recognizable images and improved caption generation performance.

\paragraph{Impact of perceptual loss in image generator}
Similar to the previous experiment, we replaced the perceptual loss in image generator (Eq.~\eqref{eqn:obj_pixel_generator}) with the $\ell_1$ 
loss on pixels. 
The fourth row in Table~\ref{tab:ablation_study} summarizes the results.
As it shows, employing perceptual loss in the image generator critically improves the performance by reducing visual differences between real and synthesized images.

\paragraph{Impact of attention in image generator}
Our image generator combines features from the text embedding $\mathbf s$ and semantic layout $\mathbf M$ by attention mechanism.
To see its impact on text-conditional image generation, we remove the attention mechanism from the image generator (computation of $\mathbf S$ and $\mathbf A^g$ in Figure~\ref{fig:image_generator_detail}) and concatenate the layout feature $\mathbf A$ directly to text embedding. 
As shown in the last row of Table~\ref{tab:ablation_study}, employing attention mechanism improves the text-conditional image generation performance, since it forces the model to exploit text information in generation process. 
We found that the attention mechanism helps the model to generate textures and background relevant to the input text.

\bigskip
\subsection{More qualitative examples}

\paragraph{Image and layout generation.}
We present the end-to-end image generation results of our method in Figure~\ref{fig:samples_box_mask_image_supp}, including object bounding boxes and masks obtained by the layout generator. 
As illustrated in the figure, our model generates object bounding boxes that match content of the input text, and shapes capturing class-specific visual attributes and relation with other objects (\eg person riding a motorcycle, person swinging a bat, etc).
Given the layout, the image generator correctly predicts object textures and background match the description.

\begin{figure*}[t!]
\centering
\includegraphics[width=0.499\linewidth]{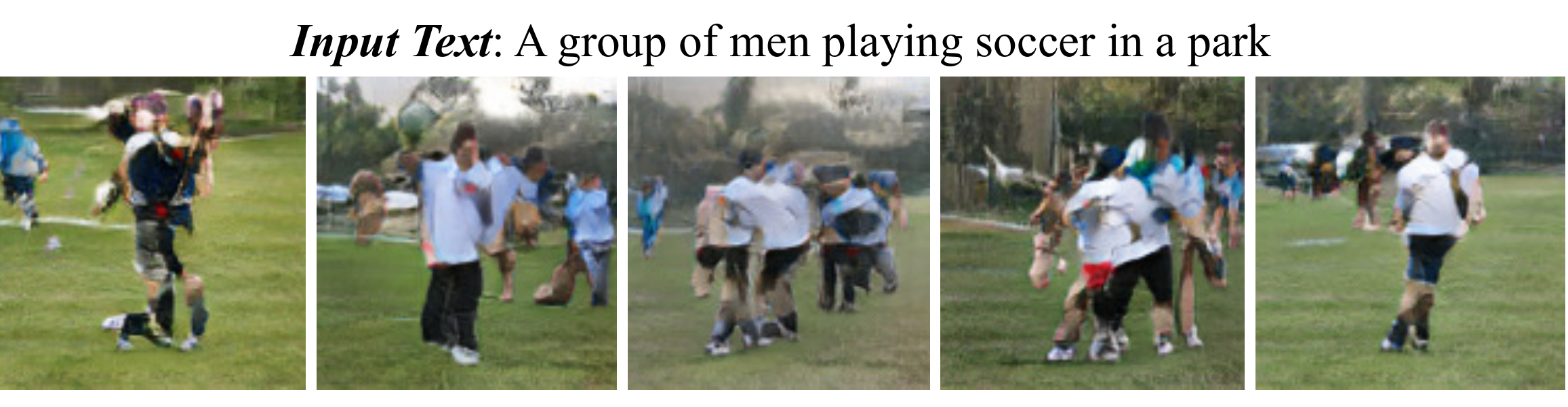}~ 
\includegraphics[width=0.499\linewidth]{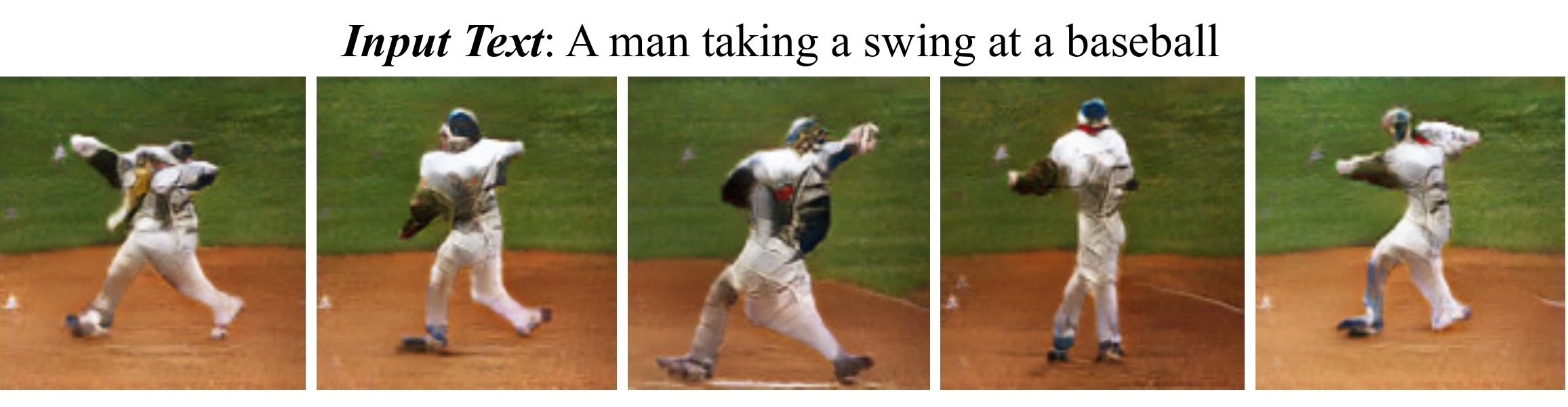} \\
\includegraphics[width=0.499\linewidth]{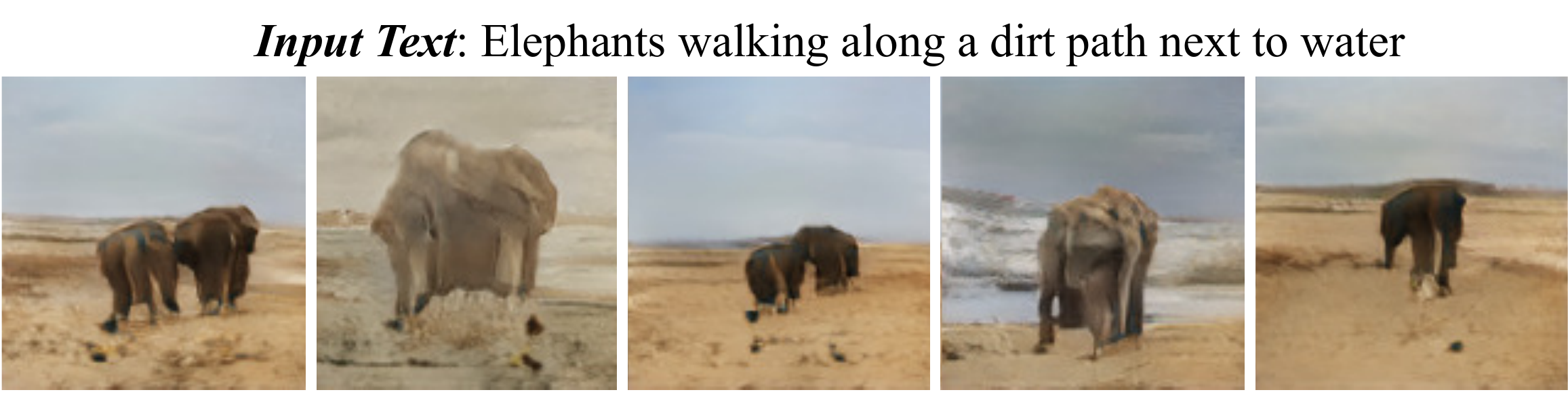}~ 
\includegraphics[width=0.499\linewidth]{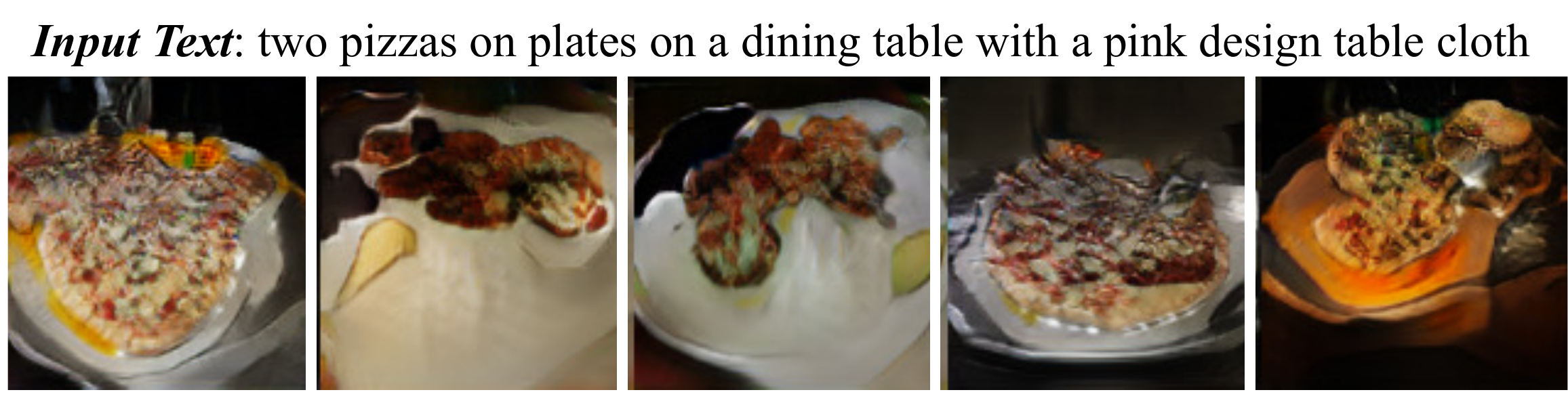} \\
\includegraphics[width=0.499\linewidth]{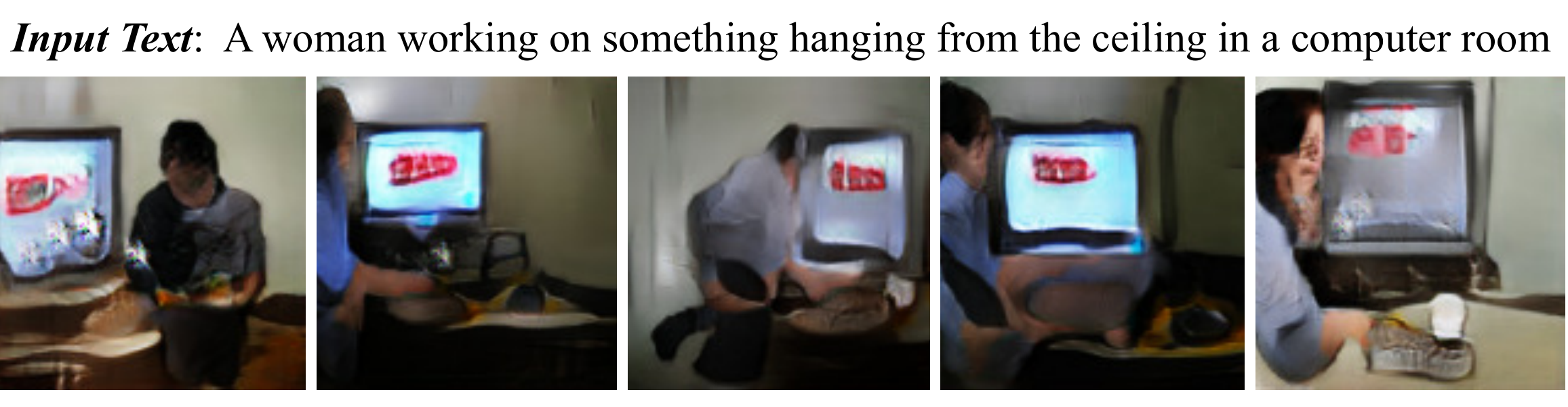}~ 
\includegraphics[width=0.499\linewidth]{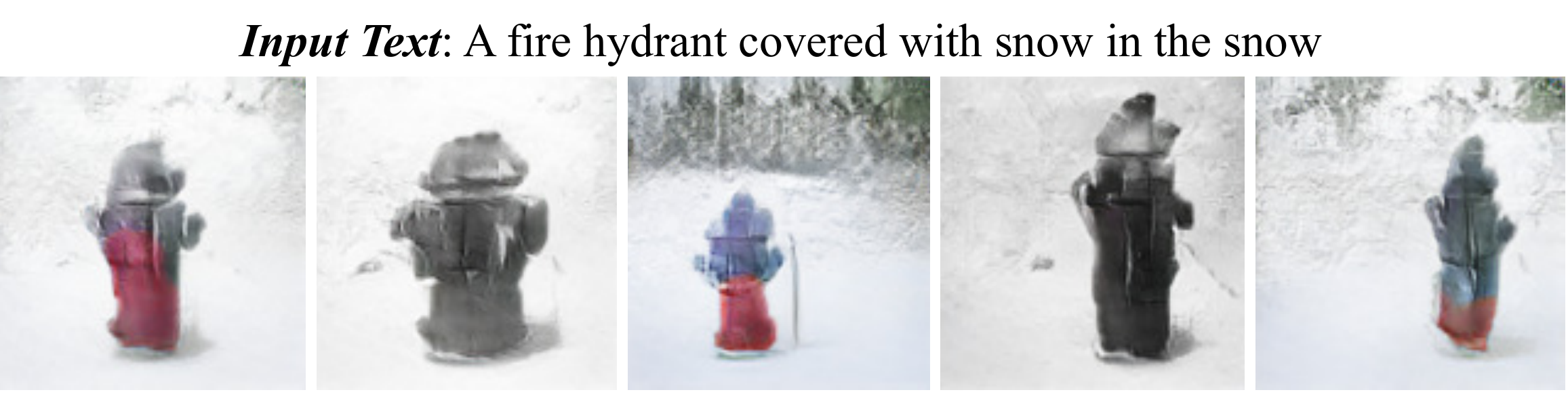} \\
\includegraphics[width=0.499\linewidth]{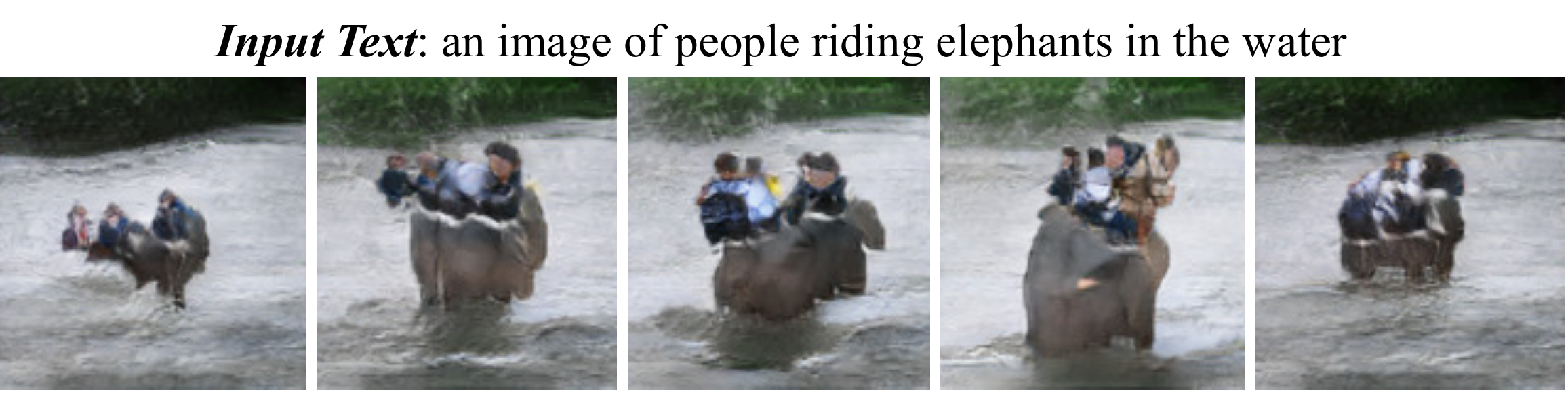}~ 
\includegraphics[width=0.499\linewidth]{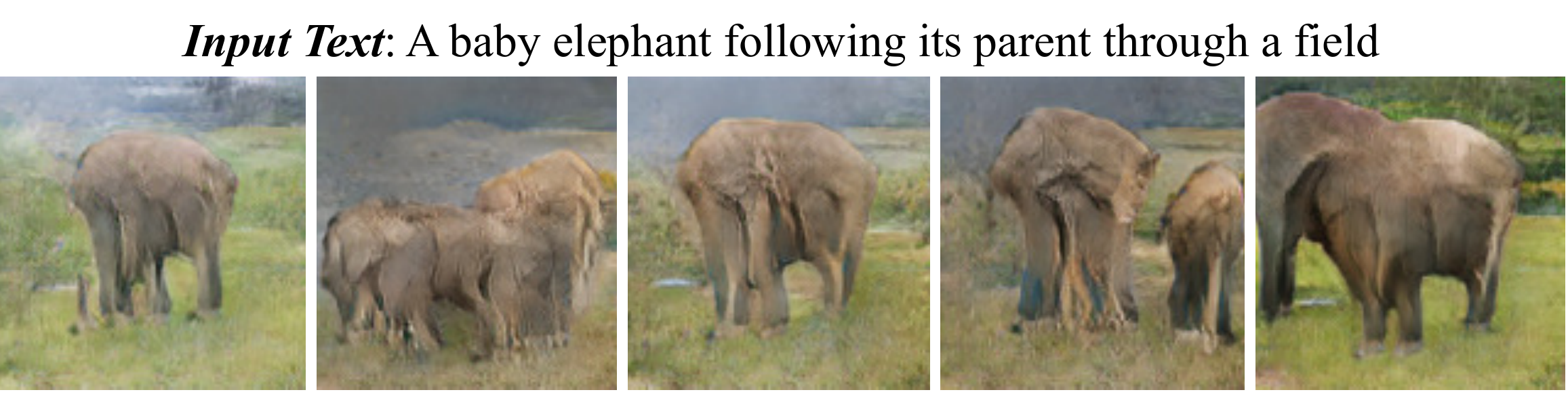} \\
\includegraphics[width=0.499\linewidth]{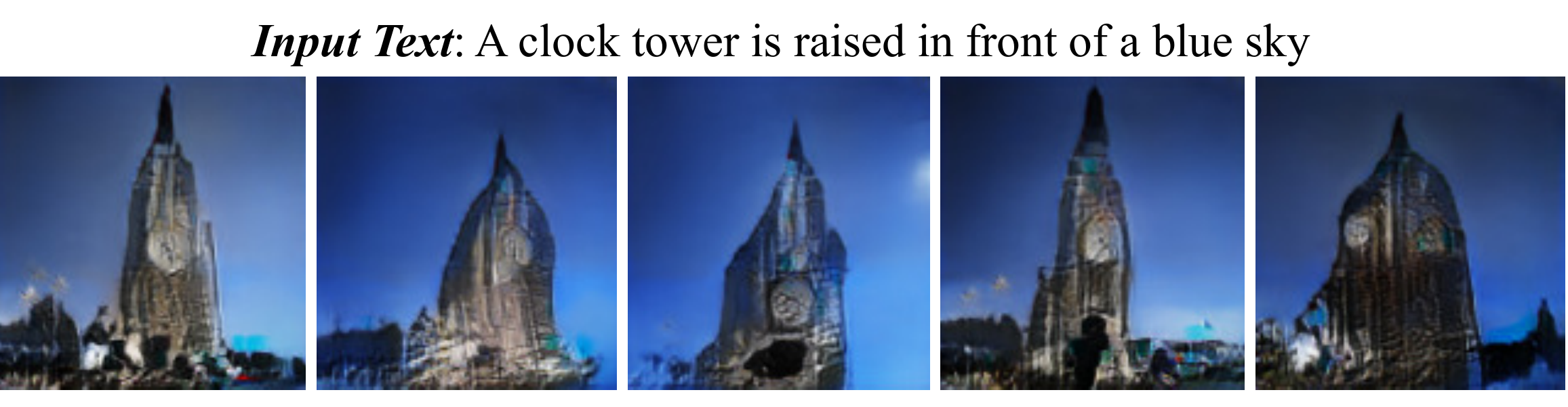}~ 
\includegraphics[width=0.499\linewidth]{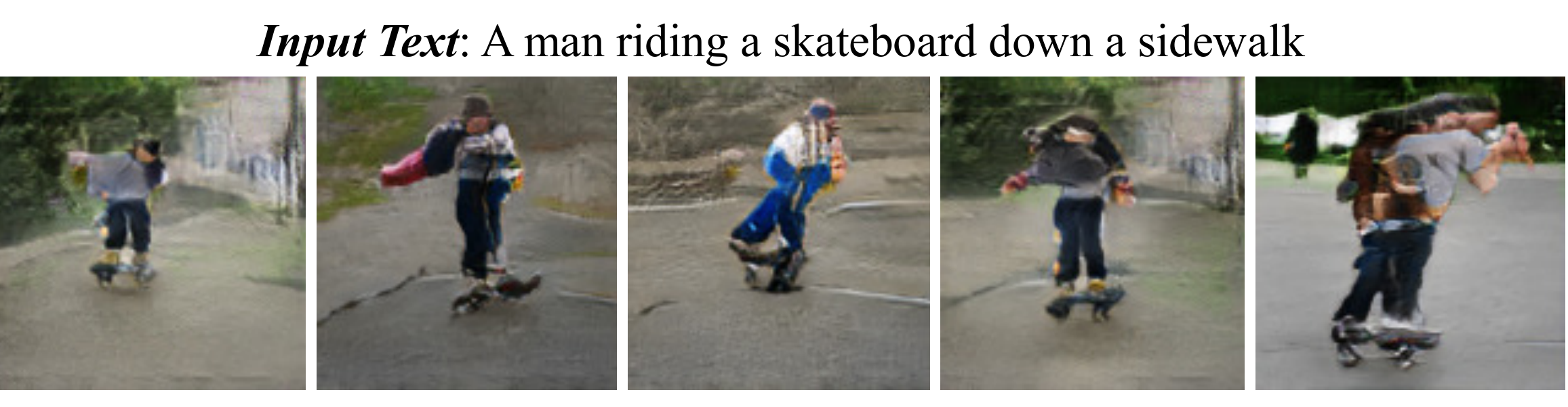} \\
\includegraphics[width=0.499\linewidth]{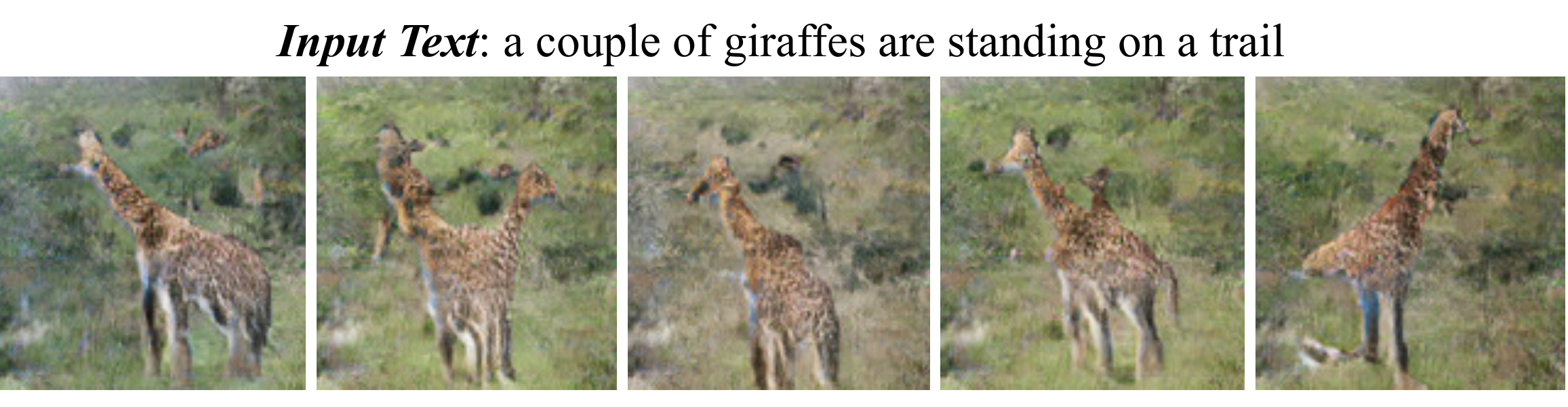}~ 
\includegraphics[width=0.499\linewidth]{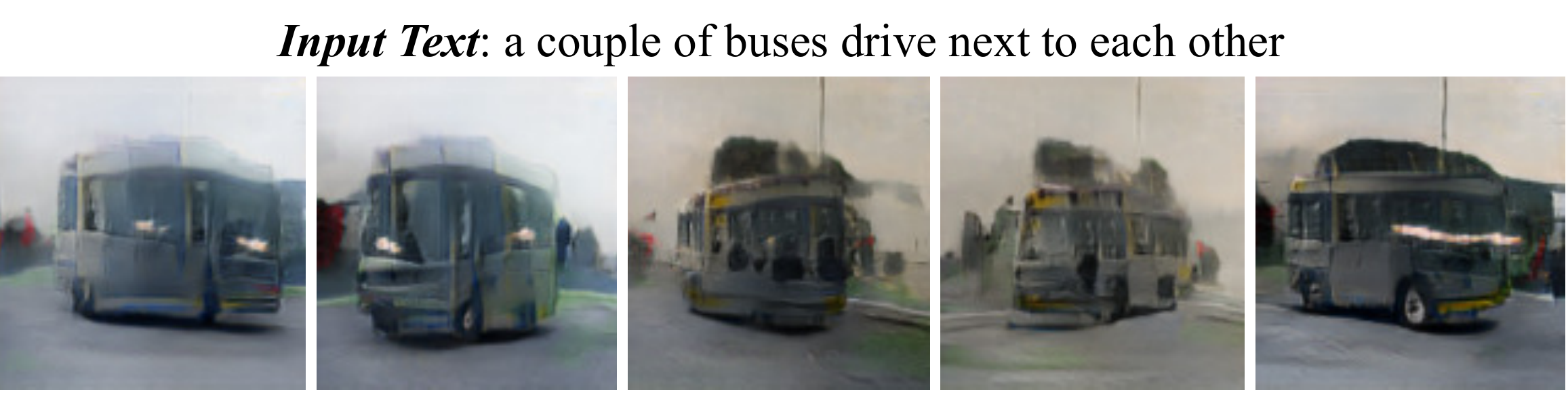} \\
\includegraphics[width=0.499\linewidth]{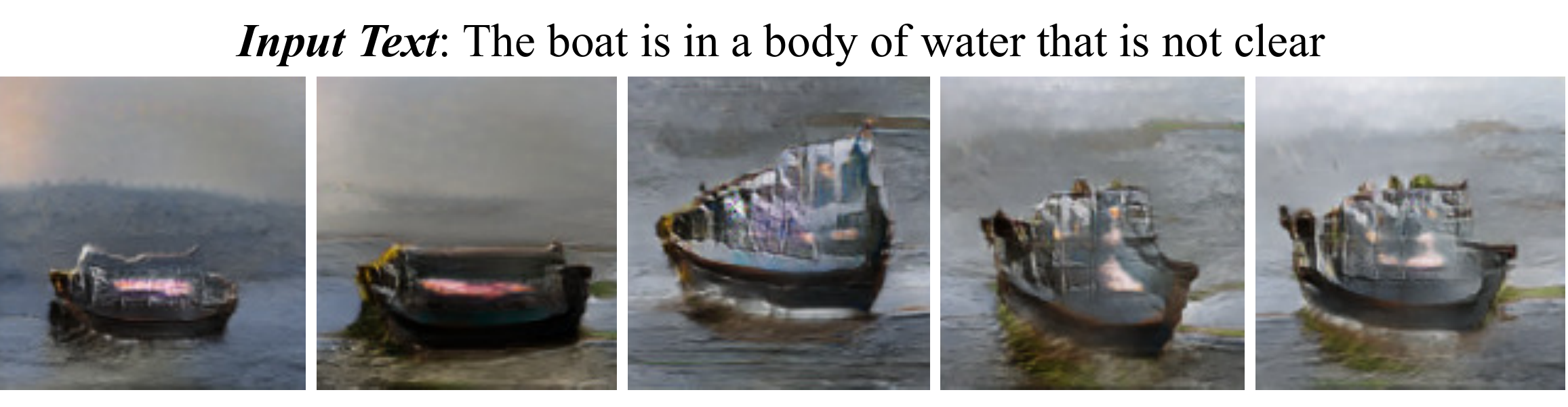}~ 
\includegraphics[width=0.499\linewidth]{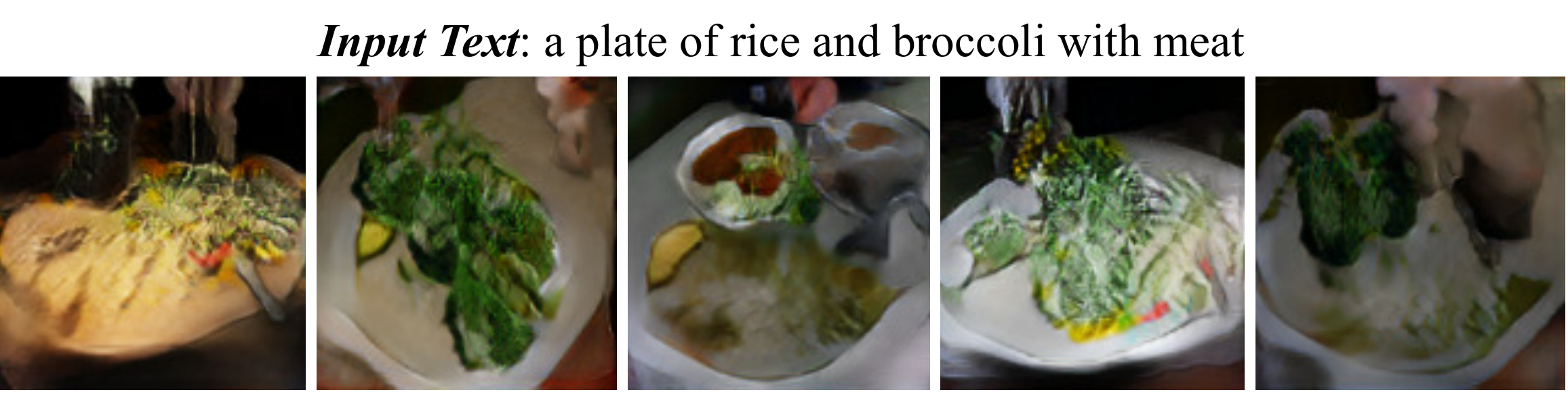} \\
\includegraphics[width=0.499\linewidth]{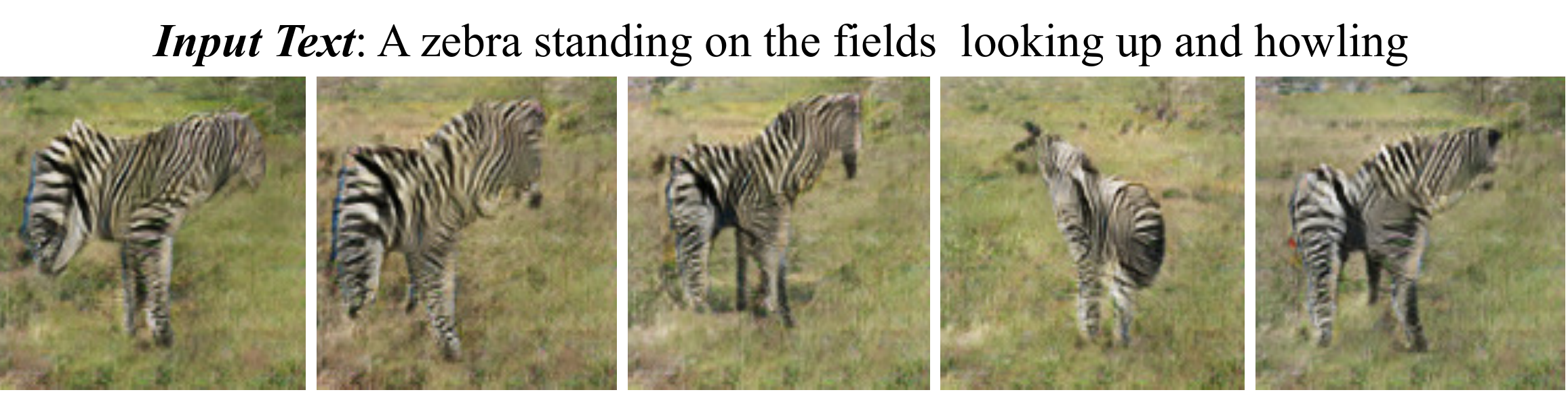}~ 
\includegraphics[width=0.499\linewidth]{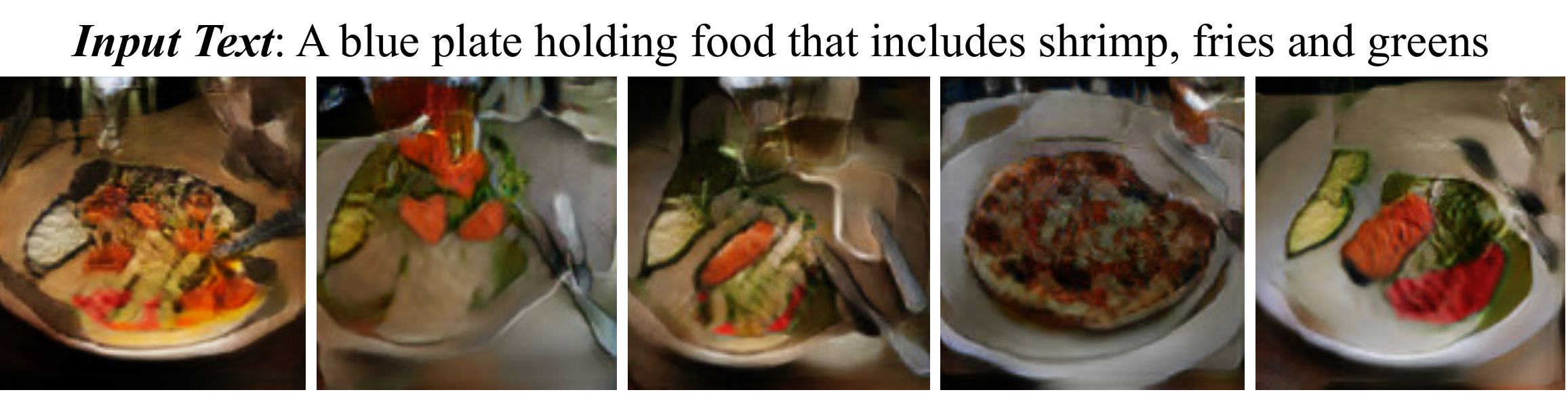} \\
\caption{\small Examples of multiple samples generated from a text description.}
\label{fig:diversity_supp}
\end{figure*}

\paragraph{Diversity of samples.}
Figure~\ref{fig:diversity_supp} presents a set of samples generated by our method, which corresponds to Figure~\ref{fig:diversity} in the main paper.
Our method generates diverse samples by generating semantic layouts that are both diverse and highly related to the input text description.

\begin{figure}[!t]
\centering
\small
\includegraphics[width=1\linewidth]{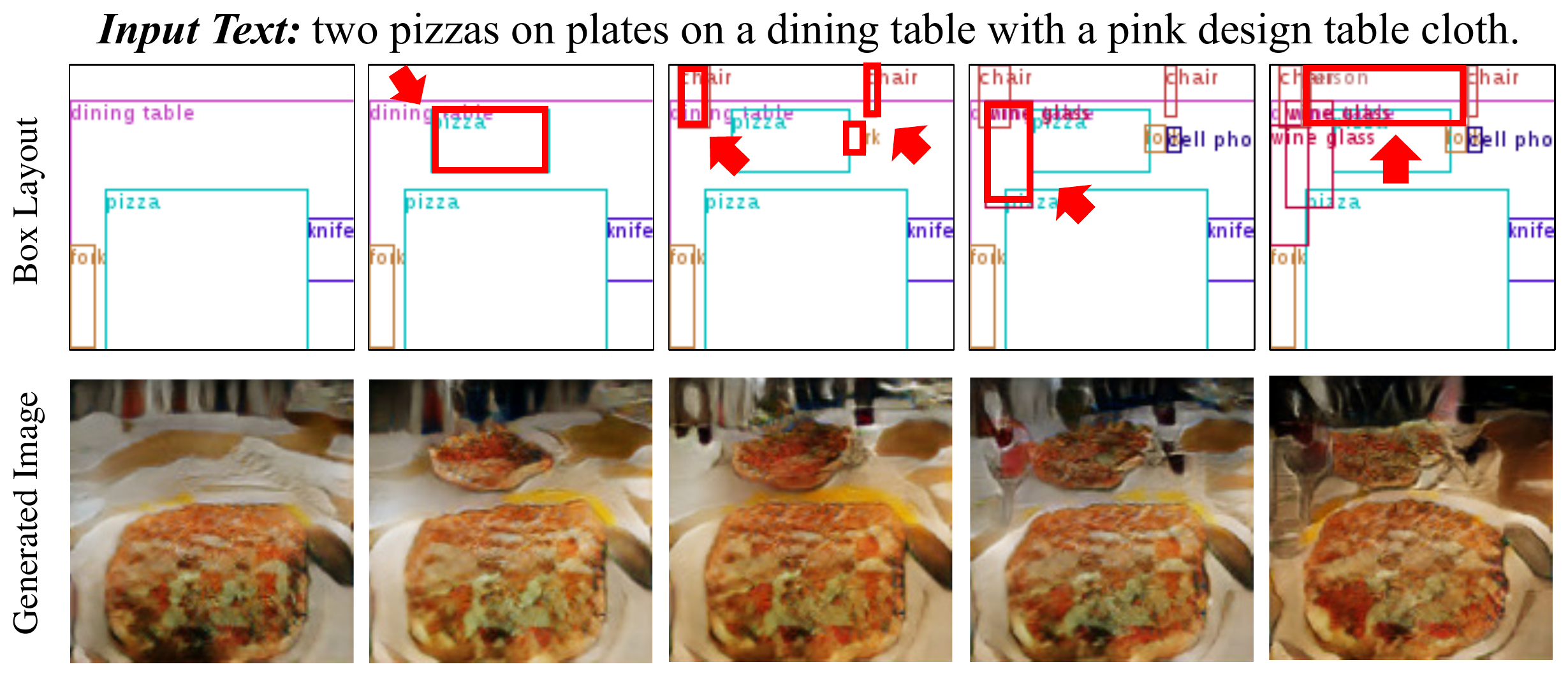} \\
\includegraphics[width=1\linewidth]{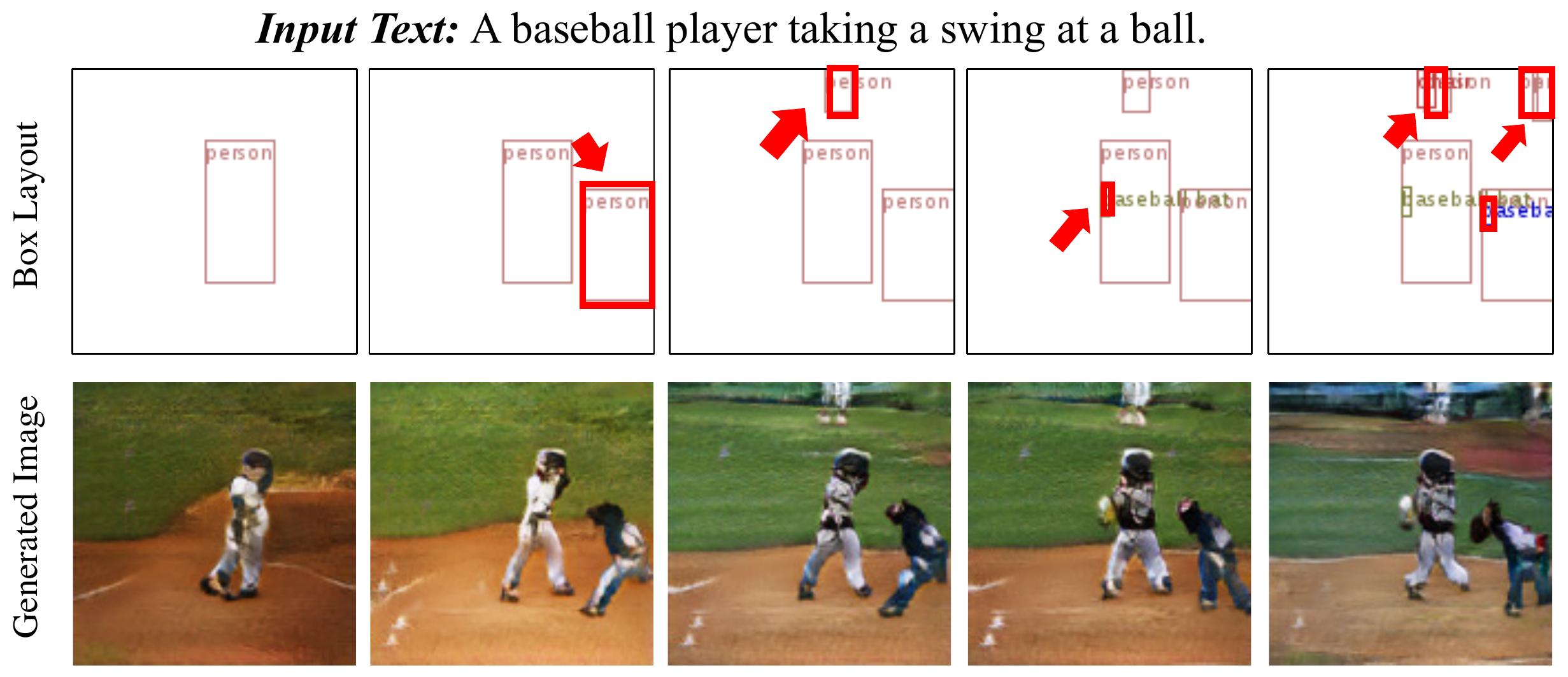} \\
\includegraphics[width=1\linewidth]{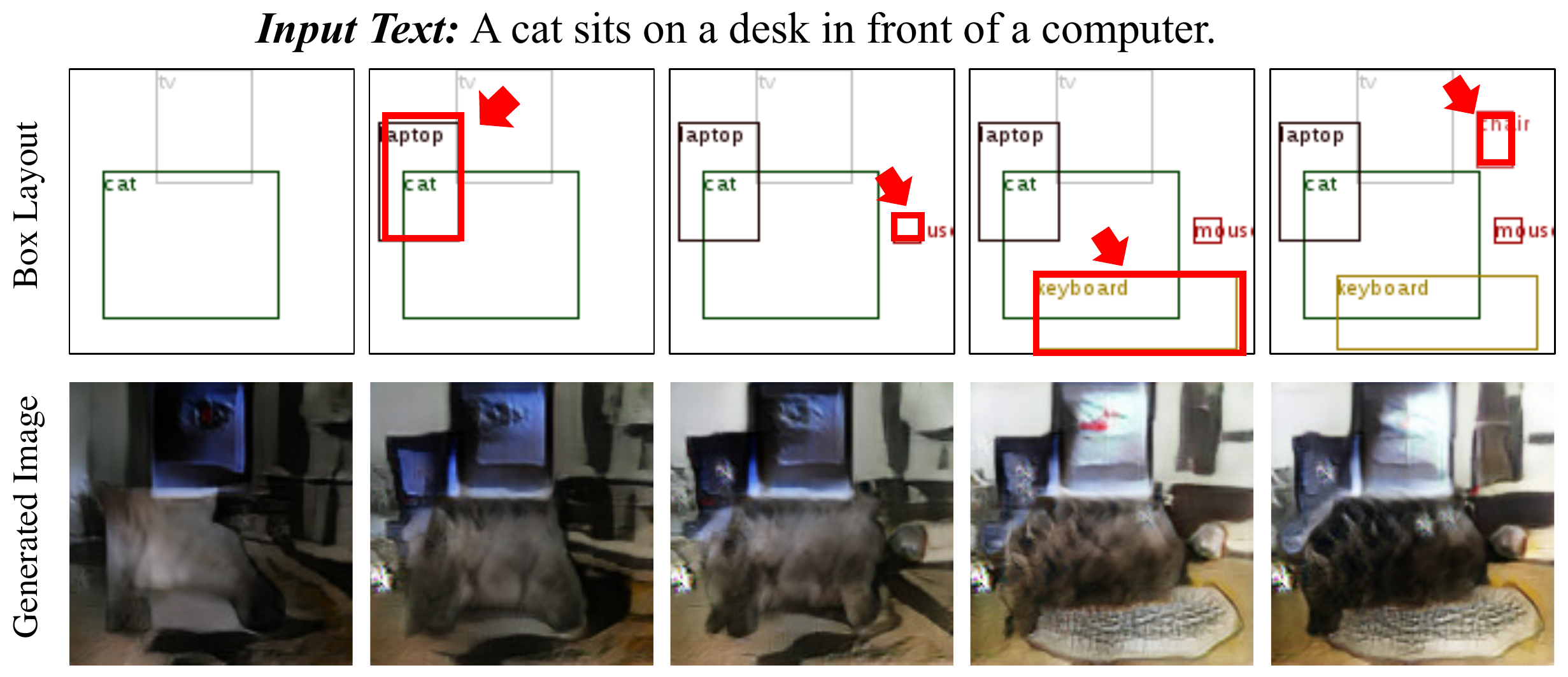} \\ 
\includegraphics[width=1\linewidth]{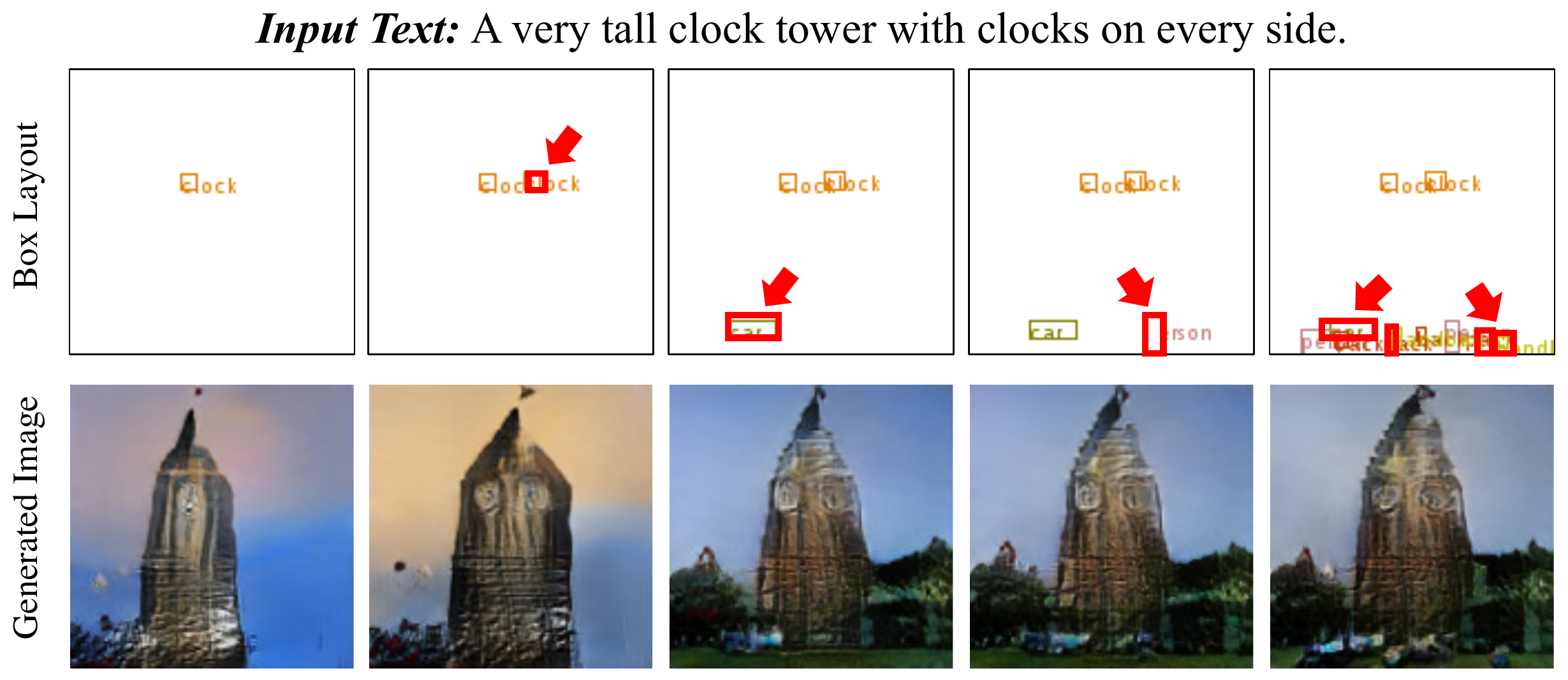} \\
\includegraphics[width=1\linewidth]{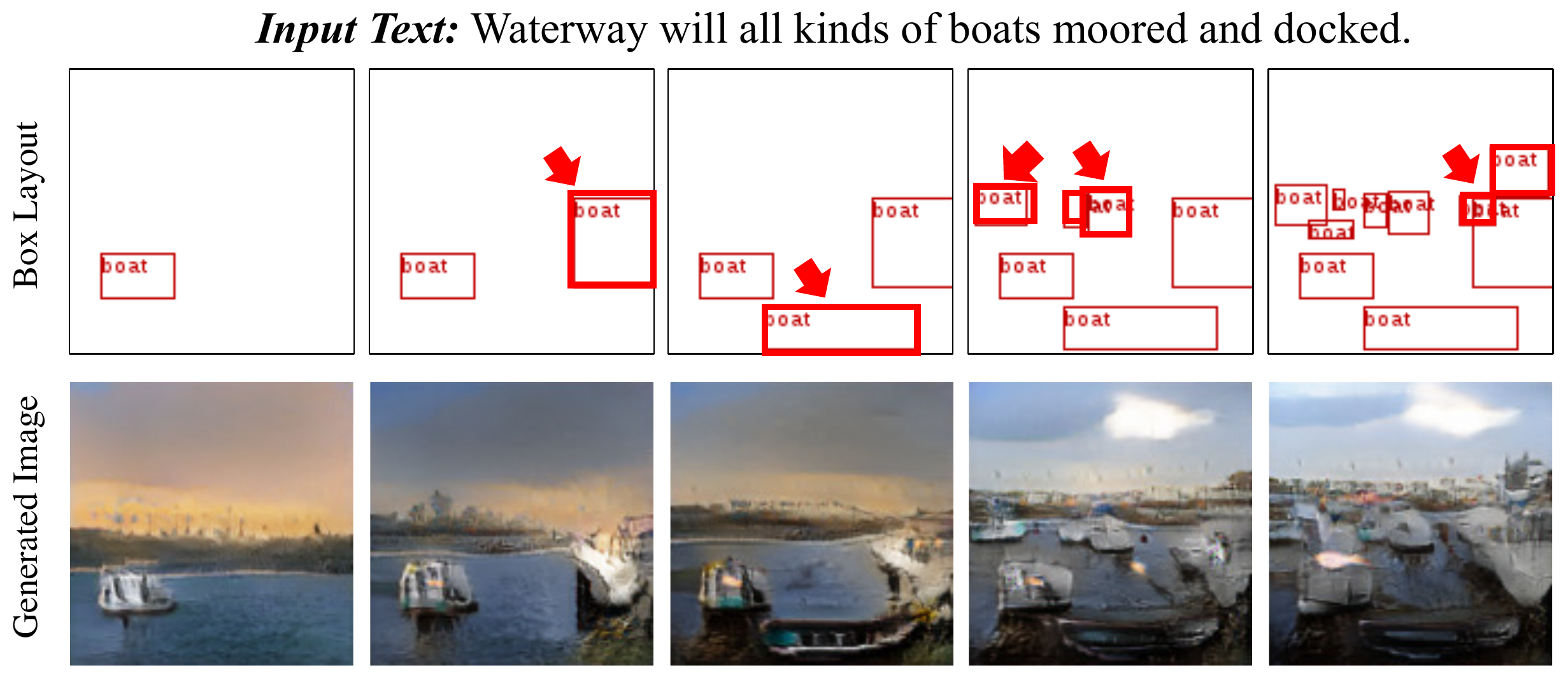} \\
\caption{
Examples of controllable image generation by adding new objects.
}
\label{fig:controllable_gen_add_supp}
\end{figure}

\begin{figure}[!t]
\centering
\small
\includegraphics[width=1\linewidth]{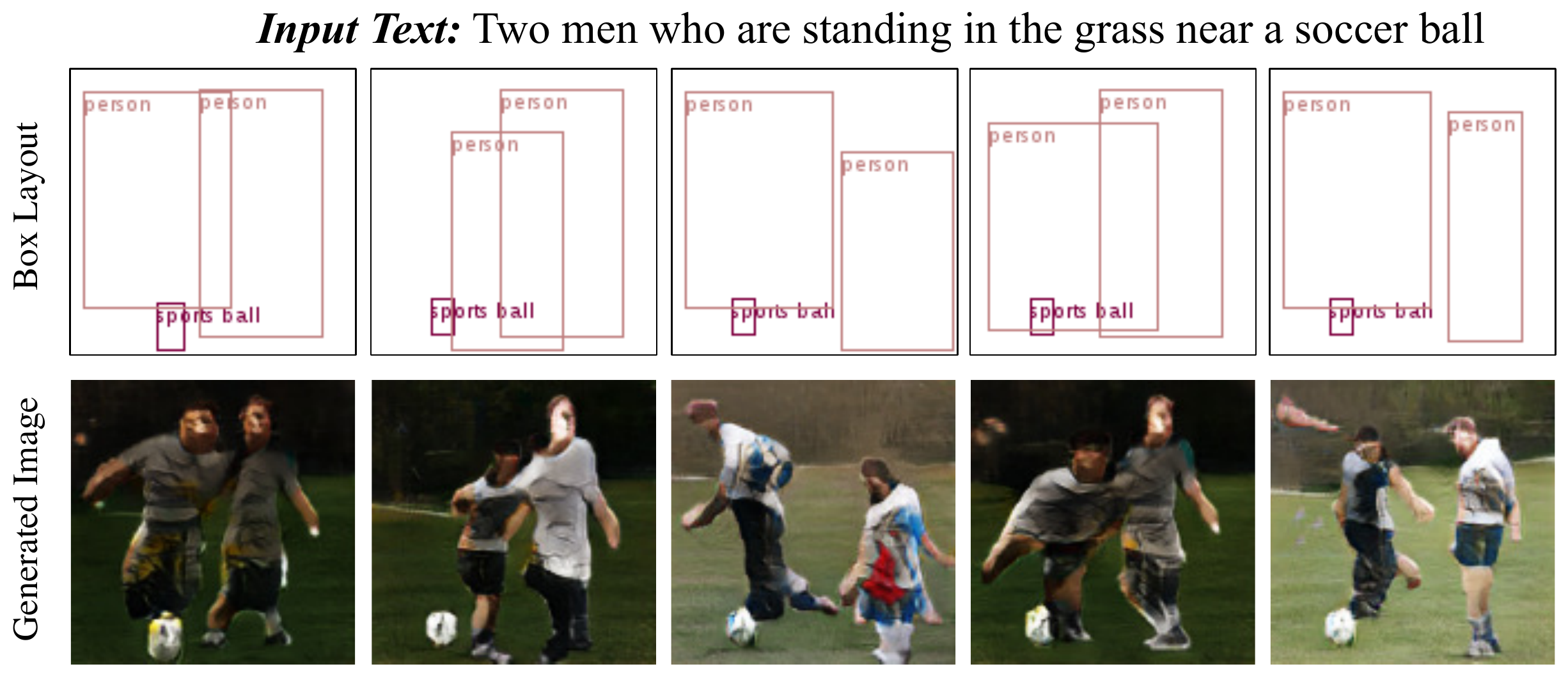} \\
\includegraphics[width=1\linewidth]{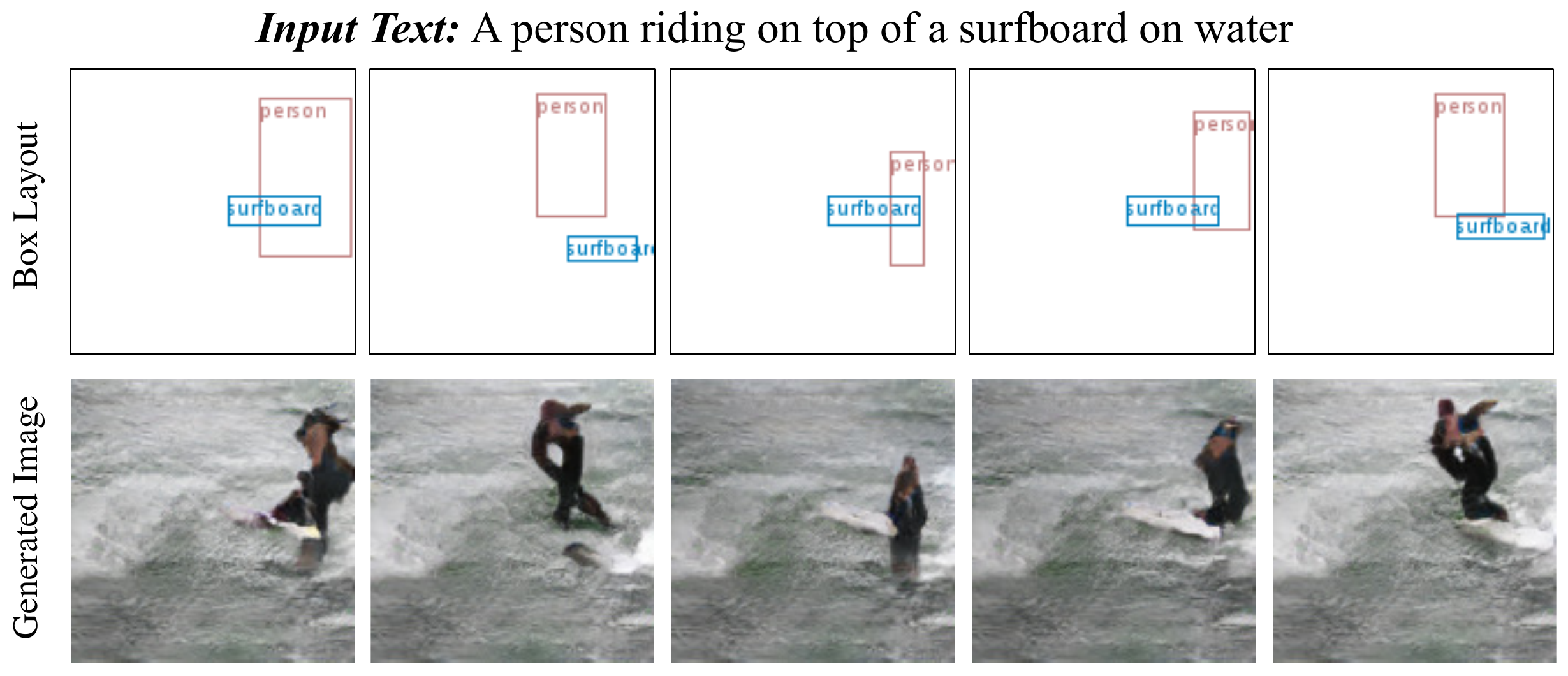} \\
\includegraphics[width=1\linewidth]{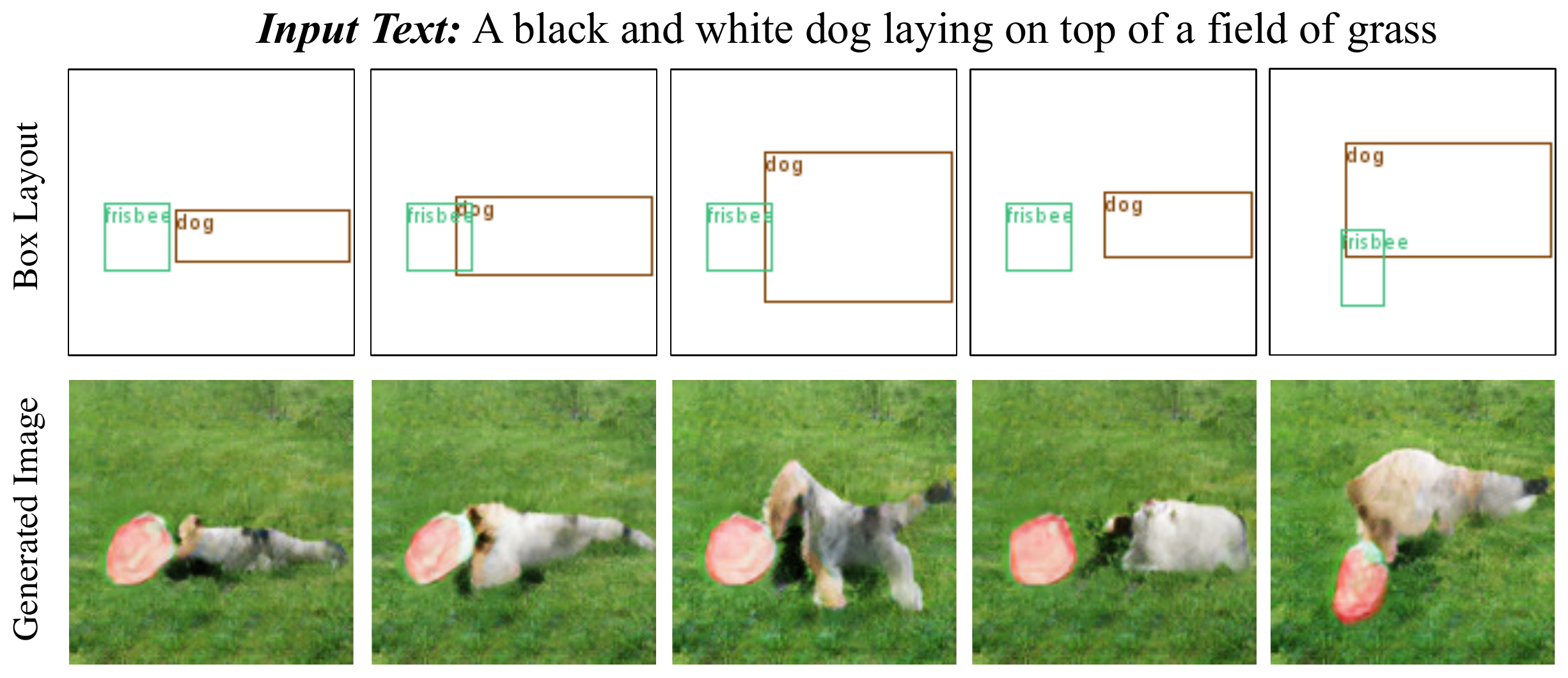} \\
\includegraphics[width=1\linewidth]{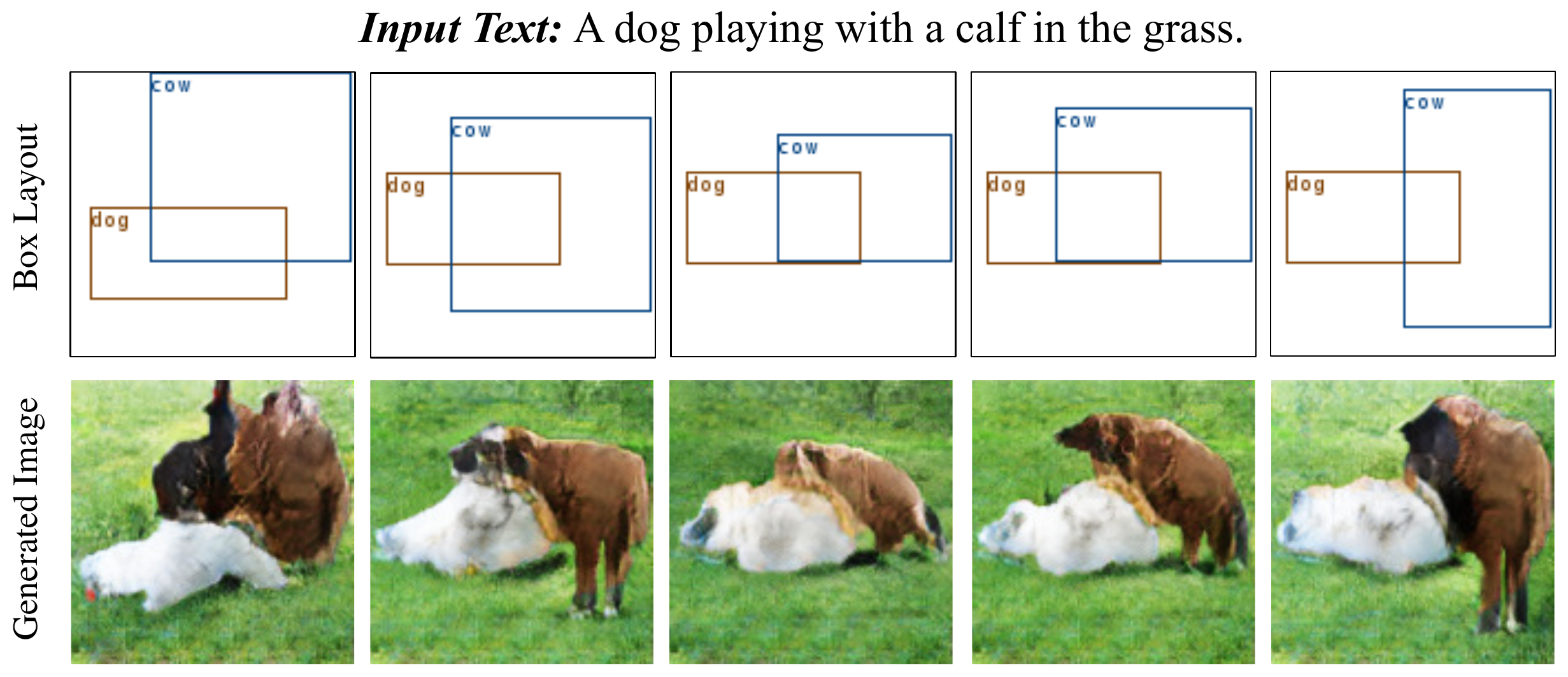} \\
\includegraphics[width=1\linewidth]{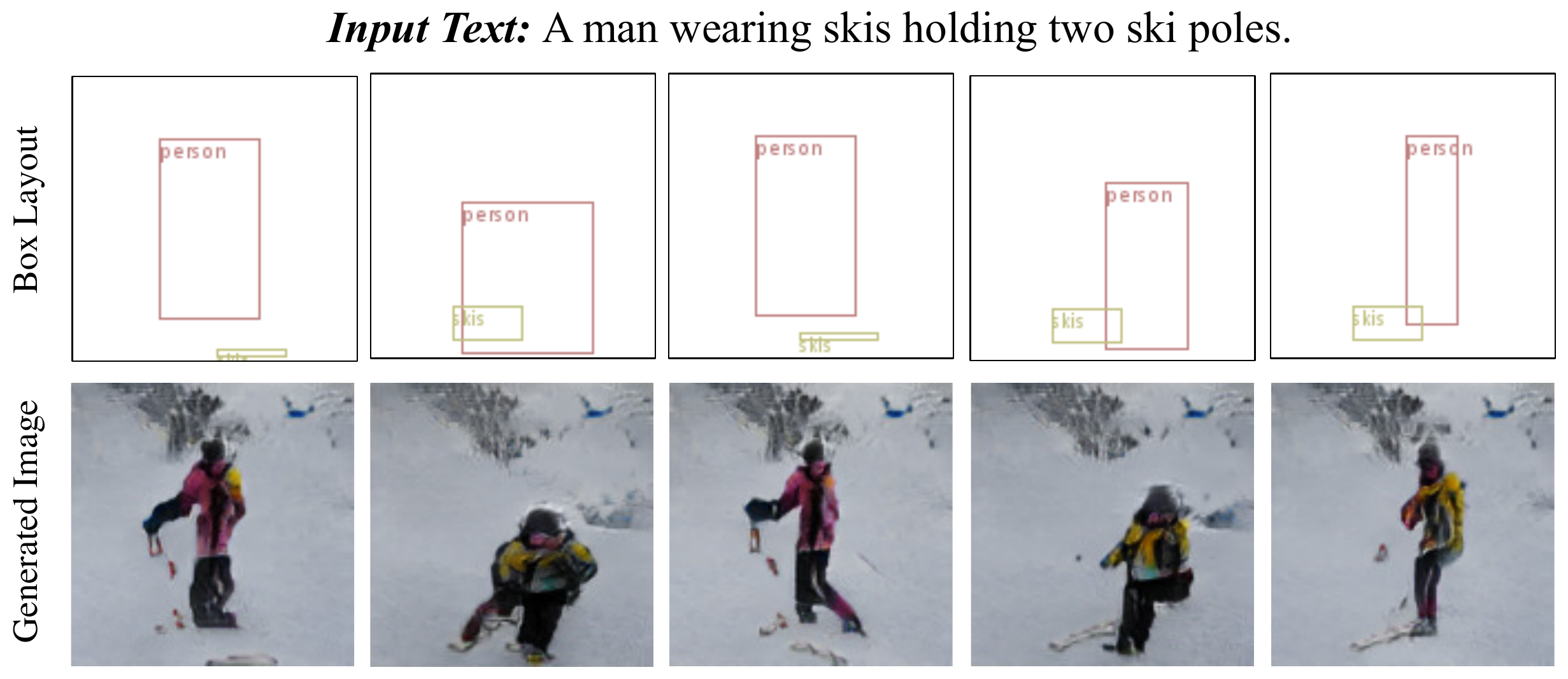} \\
\caption{
Examples of controllable image generation by modifying the size and locations of object bounding boxes.
}
\label{fig:controllable_gen_move_supp}
\end{figure}

\paragraph{Controllable image generation.}
Semantic layout provides a natural and interactive interface for image editing.
By modifying the bounding box layout of the scene, our model can generate the object shapes and images compatible with the modified layout.
Figure~\ref{fig:controllable_gen_add_supp} illustrates the generated images obtained by adding new objects to the existing semantic layout.
By placing new object bounding boxes to a scene, our model not only creates the corresponding object instance but also modifies surrounding context adaptive to the change. 
For instance, adding cars and pedestrians in front of a tower makes the model to generate a street on a background (the 4th row in Figure~\ref{fig:controllable_gen_add_supp}).
Similarly, one can modify the semantic layout by changing size and spatial location of existing objects. 
Figure~\ref{fig:controllable_gen_move_supp} illustrates the results.
Modifying the spatial configuration of objects sometimes changes the relationship between objects and leads to images in different context. 
For instance, changing the locations of a soccer ball and players leads to various images such as dribbling, shooting and competing to occupy the ball (the first row in Figure~\ref{fig:controllable_gen_move_supp}).


\end{document}